\documentclass[11pt]{article}

\usepackage[preprint]{acl}

\usepackage{times}
\usepackage{latexsym}
\usepackage{enumitem}

\usepackage[T1]{fontenc}
\usepackage{adjustbox}
\usepackage[utf8]{inputenc}

\usepackage{microtype}

\usepackage{inconsolata}

\usepackage{graphicx}
\usepackage{multirow}

\usepackage{amsmath}
\usepackage{booktabs}
\usepackage{siunitx}
\usepackage{subcaption}
\usepackage{amssymb} 
\usepackage[most]{tcolorbox}
\usepackage{soul,xcolor}

\newcommand{\grn}[2][green]{%
  \sethlcolor{#1}\hl{#2}%
}

\newcommand{\grayit}[1]{\textit{\textcolor{gray}{#1}}}

\title{Text as a Universal Interface for Transferable Personalization}

\author{
    \textbf{Yuting Liu\textsuperscript{1,2}}\thanks{Equal contribution.},
    \textbf{Jian Guan\textsuperscript{2}}\footnotemark[1],
    \textbf{Jia-Nan Li}\textsuperscript{2,4}\\
    \textbf{Wei Wu\textsuperscript{3}}\thanks{Corresponding authors: Guibing Guo and Wei Wu.},
    \textbf{Jiang-Ming Yang\textsuperscript{3}},
    \textbf{Jianzhe Zhao\textsuperscript{1}},
    \textbf{Guibing Guo\textsuperscript{1}}\footnotemark[2]
\\
\\
    \textsuperscript{1}Software College, Northeastern University, China
    \textsuperscript{2}Ant Group
    \textsuperscript{3}Ant International \\
    \textsuperscript{4}Gaoling School of Artificial Intelligence, Renmin University of China
\\
\\
\texttt{liuyuting@stumail.neu.edu.cn},~\texttt{lijianan@ruc.edu.cn}\\
\texttt{\{jianguanthu,wuwei19850318\}@gmail.com}\\
\texttt{jmyang@ant-intl.com},~\texttt{\{guogb,zhaojz\}@swc.neu.edu.cn}\\
}

\begin{document}
\maketitle
\begin{abstract}

We study the problem of personalization in large language models (LLMs). Prior work predominantly represents user preferences as implicit, model-specific vectors or parameters, yielding opaque “black-box” profiles that are difficult to interpret and transfer across models and tasks. In contrast, we advocate natural language as a universal, model- and task-agnostic interface for preference representation. The formulation leads to interpretable and reusable preference descriptions, while naturally supporting continual evolution as new interactions are observed. To learn such representations, we introduce a two-stage training framework that combines supervised fine-tuning on high-quality synthesized data with reinforcement learning to optimize long-term utility and cross-task transferability. Based on this framework, we develop \textsc{AlignXplore+}, a universal preference reasoning model that generates textual preference summaries. Experiments on nine benchmarks show that our 8B model achieves state-of-the-art performance—outperforming substantially larger open-source models—while exhibiting strong transferability across tasks, model families, and interaction formats.

\end{abstract}

\section{Introduction}

Personalization, the tailoring of model behavior to individual user needs, is a cornerstone of modern artificial intelligence (AI) applications~\citep{kirk2023the,kasirzadeh2023conversation}, especially as large language models (LLMs) transition from general-purpose tools into user-centric products~\citep{guan2025surveypersonalizedalignment}. 
Central to this endeavor is the challenge of user representation learning: the process of distilling a user's complex preferences from their behaviors into a machine-readable format~\citep{li2021survey}. For decades, the dominant paradigm has encoded user preferences into either dense vectors such as static user embeddings~\citep{bpr,ning2024user} and dynamic sequential states~\citep{sasrec}, or user-specific parameters like LoRA~\citep{hu2022lora} and adaptors~\citep{houlsby2019adapter}. 
These approaches have driven significant advancements in a wide range of applications, such as content recommendation~\citep{tallrec,cora}, targeted advertising~\citep{tucker2014social,feizi2024onlineadvertisementsllmsopportunities}, and personalized response generation~\citep{li2016persona,zhang2018personalizing,zhang-etal-2025-personalized,chen2025popi,hyunji2025learning}.

\begin{figure*}
    \centering
    \includegraphics[width=\linewidth]{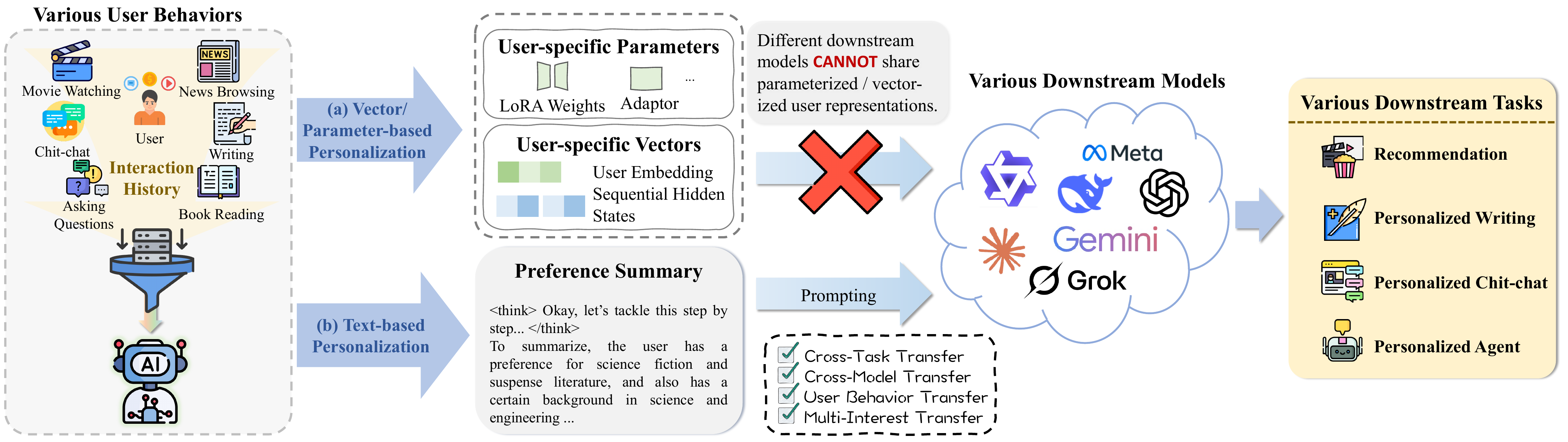}
    \caption{Paradigm shift from vector/parameter-based \textbf{(a)} to text-based user representations \textbf{(b)} for personalization.
(a) Traditional methods generate user-specific parameters and vectors that are tightly coupled with their training model, thus cannot be transferred.
(b) We pioneer a text-based paradigm that infers a model- and task-agnostic preference summary, which serves as a universal interface to empower any downstream model for various tasks.}
    \label{fig:model}
\end{figure*}
Despite their success, these representations pose fundamental barriers to a unified, user-centric AI. \textbf{First, they are opaque ``black boxes.''} This lack of human-readable meaning hinders interpretability and prevents users from understanding or controlling their own digital profiles~\citep{zhang2020explainable,charte2020analysis}. 
\textbf{Second, and more critically, they lack the zero-shot transferability required for a unified user experience.} Here, ``transferability'' signifies that the user representation itself can be fully understood by any model for any task, eliminating the need for task-specific fine-tuning or adaptation layers. However, vector or parameter-based representations are tightly coupled with their original model's architecture and objective, rendering them meaningless to a new application and creating fragmented data silos. For instance, a user's rich movie viewing history remains unknown to their music streaming service, forcing every application to learn preferences from scratch.

In this paper, we propose using text as a universal, model- and task-agnostic interface for user representation (Figure~\ref{fig:model}). Our approach synthesizes user history into a natural language preference summary. This textual representation is inherently interpretable, empowering users to inspect, edit, and control their digital profiles. It is also universally transferable; by decoupling the profile from model architecture, it enables seamless knowledge transfer across applications and the composition of a holistic user view. Furthermore, it is naturally suited for streaming inference, i.e., producing an updated summary by consuming the old one along with new interactions~\citep{alignxplore}. To generate such transferable summaries, we introduce a novel two-stage training framework, from which we derive a universal preference inference model, \textsc{AlignXplore+}. First, Supervised Fine-Tuning (SFT) bootstraps the model on high-quality data from our innovative ``generate-validate-merge'' pipeline, ensuring the initial summaries are both comprehensive and behaviorally accurate. Second, Reinforcement Learning (RL) directly optimizes the policy for transferability using two key innovations: (1) a curriculum pruning strategy that isolates reasoning-intensive signals to foster the learning of intrinsic preference patterns; and (2) a cumulative reward that promotes evolvable summaries, making them effective foundations for streaming updates of user preferences.

Extensive experiments across nine diverse benchmarks, involving recommendation, response selection, and response generation tasks, demonstrate that our \textsc{AlignXplore+} achieves state-of-the-art performance and unprecedented transferability. Our 8B model achieves an average score of 75.10\%, yielding an absolute improvement of 4.2\% over GPT-OSS-20B, and consistently outperforming all baselines, including open-source models that are orders of magnitude larger, with particularly strong gains on complex reasoning tasks. Critically, we fulfill the promise of a universal interface by demonstrating systematic transferability across several key dimensions: our textual summaries effectively transfer knowledge across diverse tasks (with up to 14.00\% absolute improvement), provide consistent gains across different model architectures (ranging from 0.76\% to 8.70\% improvements on different tasks with different downstream models), and remain robust to complex, real-world user histories, including fundamental shifts in interaction format (e.g., positive-only logs, with a 2.36\% improvement over Qwen3-8B) and the presence of multiple concurrent interests. In summary, our contributions are:

\noindent I. We pioneer a new paradigm for user representation that shifts from opaque, non-transferable vectors to interpretable, universal, and composable textual preference summary.
    
\noindent II. We introduce a novel two-stage training framework that combines SFT with RL and develop \textsc{AlignXplore+}, a reasoning model to generate transferable preference summaries.

\noindent III. Extensive experiments on diverse personalization benchmarks demonstrate that our method not only achieves state-of-the-art performance but also exhibits remarkable zero-shot transferability across downstream models, tasks and domains\footnote{Code is available at \url{https://github.com/AntResearchNLP/AlignX-Family}.}.

\section{Related works}
\subsection{User Representation Learning}

User representation learning captures user-specific context from historical behaviors to enable personalized AI systems. Traditional neural recommenders encode users as dense embeddings, initially learned by mapping user IDs to latent vectors and optimizing them via observed interactions~\citep{bpr}. To capture the dynamic, sequential patterns in user behavior, subsequent work moved beyond static IDs, leveraging Transformers~\citep{sasrec} and Graph Neural Networks (GNNs)~\citep{ngcf,lightgcn} to encode entire user histories. Further advancements incorporated multimodal item features (e.g., images) to enrich the user representations~\citep{mmgcn,idsf}. A parallel line of research bypasses explicit user embeddings, instead encoding user interactions directly into the model's full parameters~\citep{kang2023llms,li2023teach} or partial parameters~\citep{tan2024personalized,dan2024p}. However, both vector- and parameter-based representations are inherently opaque and model-specific. 
Recent methods resolve these limitations by directly prompting LLMs with users' textual histories~\citep{xu2022beyond,tallrec,Salemi2024OptimizationMF}, which presents a dual challenge: the prohibitive computational cost and latency associated with processing long user histories~\citep{yang2025r1}, and the inherent difficulty for LLMs to induce user preferences from noisy interactions~\citep{qiu2024phenomenal,alignxplore}. Our work addresses both issues by employing reinforcement learning to train a policy that incrementally transforms noisy user histories into an efficient, interpretable, and transferable textual representation.

\subsection{LLMs for Personalization}

The rapid progress of LLMs has inspired a growing body of work on leveraging their language understanding and generation capabilities for various personalization tasks. For recommendation, LLMs can serve as sequence encoders to generate user or item embeddings~\citep{onerec}, be prompted directly to produce ranked item lists~\citep{cora}, or be used for user data augmentation~\citep{llmda,negativesampling}. For personalized generation tasks, LLMs can adapt style, tone, and content to match user preferences induced from the user interactions for dialogue, summarization, and creative writing~\citep{xu2022beyond,lamp,alignx}. Besides, LLMs can be integrated into agent frameworks equipped with explicit memory modules that store and retrieve user-specific knowledge to guide multi-step personalized interactions~\citep{ldagent}. To align these models with individual user preferences, a variety of optimization strategies have been explored for vector- and parameter-based user representations, including fine-tuning user-specific parameters~\citep{psoups,park2024rlhf} and guiding LLM decoding with user-specific rewards~\citep{chen2024pad,shi2024decoding}. Our work departs from these approaches and focuses on optimizing textual user representations through curriculum RL, enabling transferable personalization at inference time.

\section{Methodology}
\label{sec:method}

\begin{figure}[!t]
    \centering
    \includegraphics[width=\linewidth]{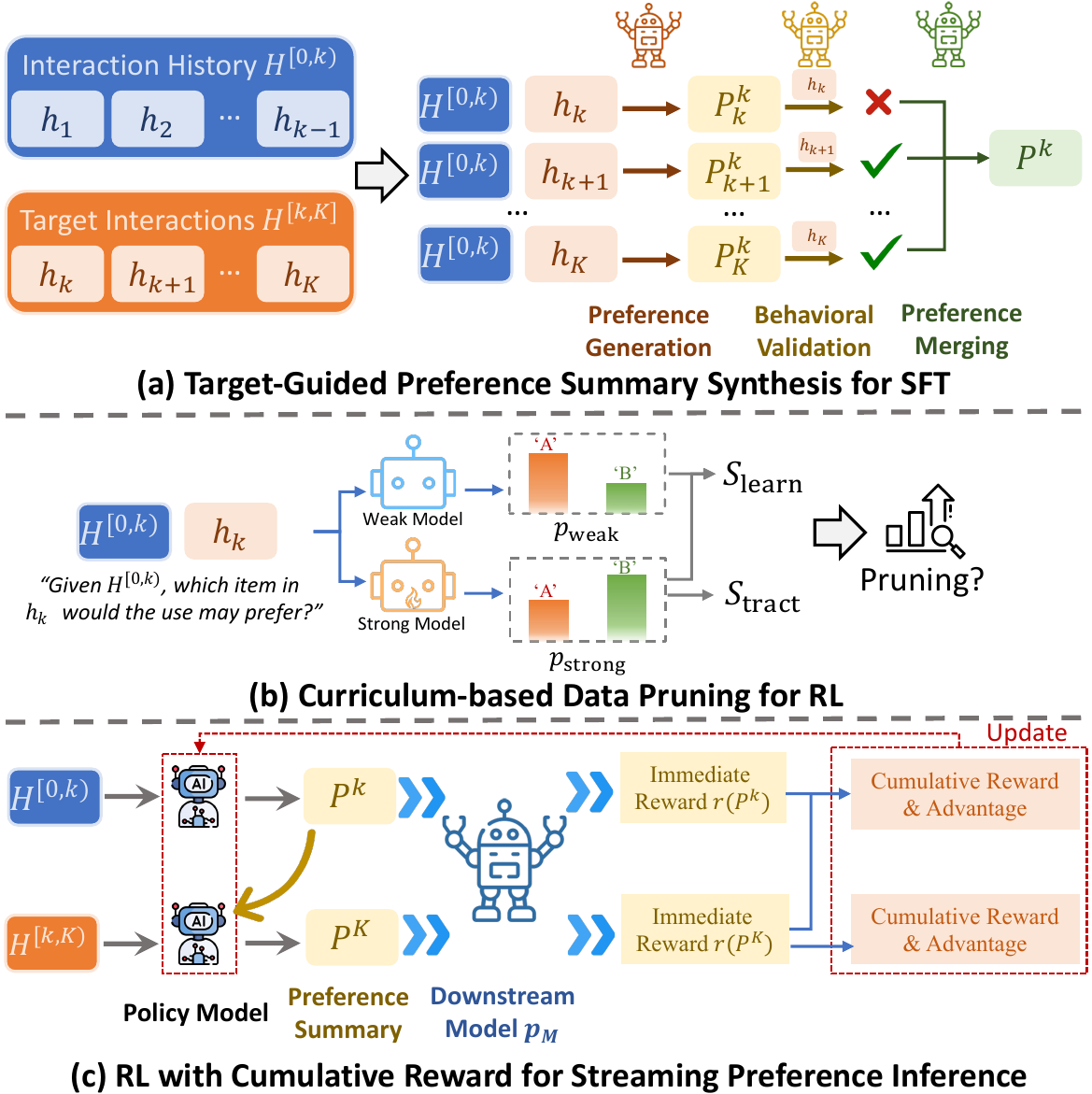}
    \caption{Overview of our training framework, which consists of two main stages. (a) SFT Stage: We first create high-quality training data using a ``generate-validate-merge'' pipeline, which synthesizes a comprehensive preference summary $P^k$ by ensuring accurate prediction of multiple future interactions. (b) \& (c) RL Stage: This stage features a curriculum pruning strategy that uses select reasoning-intensive samples, and a cumulative reward function that optimizes summaries for long-term effectiveness in streaming scenarios.}
    \label{fig:synthesis_pipeline}
\end{figure}

We aim to learn a policy that crafts a textual summary of user preferences from the user's interaction history. This policy operates incrementally: synthesizing the user's recent interactions, while also conditioning on a pre-existing preference summary~\citep{alignxplore}, which can encapsulate long-term interests or even be imported from entirely different interaction environments (e.g., from recommendation to dialogue). Our framework is thus highly efficient by obviating the need to re-process the entire interaction history at each step. To this end, we propose a novel framework combining SFT and RL, as shown in Figure~\ref{fig:synthesis_pipeline}.

\subsection{Task Formulation}

To unify diverse interaction environments, we formulate user interaction as a sequence of preference pairs. Specifically, a user's interaction history within a certain period is defined as a sequence $H^{[0,K]} = (h_0, h_1, \dots, h_K)$. Each $h_i$ is a triple $(p_i, n_i, c_i)$ where $p_i$ and $n_i$ represent a preference pair where the user prefers item $p_i$ over item $n_i$, denoted as $p_i\succ n_i$, and $c_i$ is the context in which this preference was expressed. This formulation intentionally accommodates various task modalities. For example, in recommendation systems, $c_i$ is often null, and $p_i$ could be an item the user clicked on, while $n_i$ could be an item that was displayed but ignored, or even empty if only positive feedback is available. In personalized dialogue systems, $c_i$ would be the user's preceding post or the dialogue history, and $p_i$ could be a model response that the user ``liked'' or continued the conversation from, while $n_i$ is a response they ``disliked.''

We aim to learn a generative policy $\pi_{\theta}$, parameterized by $\theta$, that incrementally crafts a textual summary of a user's preferences. Faced with a continuous stream of interactions, this policy is designed to operate on any segment of the user's history and evolve the preference summary over time. Formally, given any two time steps $k_1 < k_2$, it synthesizes an updated summary $P^{k_2}$ by conditioning on the prior summary $P^{k_1}$ and the intervening interactions $H^{[k_1, k_2)}$ through a reasoning process~\citep{openai2024o1preview,deepseekr1}:
\begin{equation} \label{eq:policy_definition}
    P^{k_{2}} \sim \pi_{\theta}\left(\cdot \mid P^{k_1}, H^{[k_1, k_{2})}\right).
\end{equation}
While RL is powerful for optimizing our policy, it may require a strong starting point. To address this, we first perform SFT to provide the policy with a strong initialization.

\subsection{Supervised Fine-tuning}
\label{sec:cold-start}

A conventional approach to synthesizing SFT data might involve prompting an off-the-shelf LLM to generate a summary directly from a user's history~\citep{alignxplore}. However, we find this unconstrained generation often produces summaries that are either too generic or fixated on a single aspect of user preference. Such summaries fail to capture the multi-faceted nature of true user preferences, severely limiting their transferability.

Our key insight is that to generate deep and comprehensive summaries, the model must be guided to explain why a user makes specific future choices. This motivates our ``divide-and-conquer'' synthesis pipeline, which uses future interactions as explicit grounding targets. As illustrated in Figure~\ref{fig:synthesis_pipeline} (a), this pipeline first generates and validates multiple, fine-grained summaries anchored to specific future behaviors, before merging them into a single holistic summary (see Appendix~\ref{app:sft_synthesis} for details). This ensures the final summary is both accurate (enabling user-aligned actions) and comprehensive (capturing the full spectrum of preferences). We present the details as follows:

\paragraph{Stage 1: Preference Generation.} To generate a summary for a given history prefix $H^{[0,k)}$, we use subsequent interactions as relevance anchors. Specifically, we sample multiple preference pairs, which we call ``target interactions'', from $H^{[k,K]}$. Each target interaction, along with the prefix $H^{[0,k)}$, forms a distinct input for an LLM to synthesize a preference summary explaining the behavior in that target. This compels the LLM to extract only the historical preferences pertinent to that specific decision, avoiding a generic and often unhelpful overview. Crucially, we treat these pairs as unlabeled targets, providing only the context and candidate items without revealing the user's actual choice, which prevents shortcut learning. 

\paragraph{Stage 2: Behavioral Validation.} To validate the quality of each generated preference summary, we use it to condition an LLM to perform preference prediction on its corresponding target interaction.  (i.e., asking ``Given this preference, which item would the user prefer?''). We discard any summary that fails to predict the user's actual choice, thereby filtering out inaccurate or inconsistent ones.
    
\paragraph{Stage 3: Preference Merging.}
After filtering, we obtain a set of target-specific preference summaries. We then prompt a powerful LLM to merge these candidates into a single, comprehensive summary. The prompt explicitly instructs the model to synthesize a non-redundant and holistic view of the user's preferences, complete with a reasoning process. Finally, to guarantee its behavioral accuracy, this merged summary undergoes a final validation, where we verify its ability to correctly predict the majority of the original target interactions.

We can naturally adapt our synthesis pipeline for the streaming setting to capture preference evolution. First, we synthesize an initial summary from an early segment of user history. Then, for each subsequent interaction segment, we synthesize an updated summary by conditioning our pipeline on both the new interactions and the previously generated summary. This iterative process creates an SFT dataset with examples of both generating a summary from scratch and incrementally updating an existing one with new user interactions.

\subsection{Reinforcement Learning}

We further refine the policy $\pi_{\theta}$ using an RL framework tailored to address two fundamental challenges in building a universal personalization model: \textbf{(1) Inefficient learning on heterogeneous data:} Raw interaction data presents a wide spectrum of difficulty. Naive training is inefficient, wasting resources on trivial instances while failing to learn from prohibitively hard ones. This biases the model towards simplistic solutions and harms its multi-task transferability~\citep{multirl}. \textbf{(2) Myopic optimization in streaming settings:} In a streaming context, a summary must be a good foundation for future updates, not just immediately accurate. A reward function focused solely on immediate utility encourages myopic, short-sighted summaries that degrade over time. 

To tackle these challenges, we introduce two corresponding innovations. For the first, we employ a ``curriculum-based data pruning'' strategy to automatically select a calibrated set of reasoning-intensive instances. For the second, we design a ``cumulative reward'' function that explicitly optimizes for the long-term, streaming utility of a summary.

\paragraph{Curriculum-based Data Pruning.}
\label{sec:data-pruning}
This strategy aims to select challenging yet tractable examples across tasks that may vary dramatically in difficulty. Formally, the full interaction history of a user, $H^{[0,K]}$, can be decomposed into a set of training instances, where each instance is a pair $(H^{[0,k)}, h_k)$ representing the task of predicting interaction $h_k$ from its preceding history. 
Our strategy employs a pair of models: a general-purpose ``weak'' model and a specialized ``strong'' model created by fine-tuning the weak one on a direct prediction task. These models are used to assess each potential training sample based on two key metrics:
\begin{itemize}[leftmargin=2em]
    \item \textbf{Tractability Score ($S_{\text{tract}}$):} Derived from the strong model's performance on a given sample $(H^{[0,k)}, h_k)$: 
    $S_{\text{tract}} = p_{\text{strong}}(p_k \succ n_k | H^{[0,k)}, h_k)$. 
    It filters out samples that are either trivially easy or impossibly hard by enforcing the condition $\tau_{\text{low}} < S_{\text{tract}} < \tau_{\text{high}}$.

    \item \textbf{Learning Potential Score ($S_{\text{learn}}$):} This measures the performance lift as the difference in log-probabilities between the strong and weak models: 
    $S_{\text{learn}} = \log \frac{p_{\text{strong}}(p_k \succ n_k | H^{[0,k)}, h_k)}{p_{\text{weak}}(p_k \succ n_k | H^{[0,k)}, h_k)}$. 
    We select for samples where this gap is in the top $\alpha$-percentile, as this effectively filters out simpler instances where the weak model already performs well.
\end{itemize}

Based on this pruned, high-signal dataset, we focus the learning process on instances where reasoning provides the greatest marginal utility. Appendix~\ref{app:pruning_details} shows more implementation details.

\paragraph{Rollout Generation.}
To explicitly train for the streaming setting, our rollout generation is designed to mimic the core dynamic of summary evolution.  For each user history, our pruning method identifies two interaction time steps $k_1 < k_2$, which define the structure for our streaming rollout generation: an initial history segment $H^{[0,k_1)}$ and a subsequent segment $H^{[k_1,k_2)}$. The policy $\pi_{\text{old}}$ then performs a two-stage hierarchical sampling process: it first generates a set of $G$ initial preference summaries from the early history, and then, conditioned on one randomly selected initial summary, generates $G$ updated summaries using the later history. This process is formalized as:
\begin{align}
    \label{eq:rollout}
    & \{P_{i}^{k_1}\}_{i=1}^G \sim \pi_{\text{old}}(\cdot | H^{[0,k_1)}), \\
    & \{P_{j}^{k_2}\}_{j=1}^G \sim \pi_{\text{old}}(\cdot \mid \hat{P}^{k_1}, H^{[k_1,k_2)}),
\end{align}
where $\hat{P}^{k_1}$ is randomly sampled from the generated summaries in the first step. This process yields a total of $2G$ candidate summaries, forming a tree-like structure of potential reasoning paths.

\paragraph{Reward and Advantage.}
Our reward function is designed to credit summaries that are not only immediately useful but also serve as a good foundation for future updates. To this end, we define an \emph{immediate reward}, $r(P)$, which repurposes the binary filter from our SFT behavioral validation (Section \ref{sec:cold-start}) into a continuous signal. This reward is the probability assigned by a powerful downstream model $M$ to the user's actual choice, given a summary $P$ and its corresponding ground-truth interaction $h=(p, n, c)$:
\begin{equation} \label{eq:immediate_reward}
    r(P) = p_{M}\big(p \succ n \mid P, c\big).
\end{equation}
Based on this, we compute a \emph{cumulative reward}, $R(P)$, that incorporates long-horizon utility. For an updated summary $P_{j}^{k_2}$ from the second rollout set, its cumulative reward is simply its immediate reward. For an initial summary $P_{i}^{k_1}$, its cumulative reward is calculated as: 
\begin{equation} \label{eq:final_reward}
    R(P_{i}^{k_1}) = r(P_{i}^{k_1}) + \gamma \cdot \frac{1}{G} \sum_{j=1}^{G} R_{j}^{k_2},
\end{equation}
where $\gamma \in [0, 1]$ is a discount factor that balances immediate accuracy with future utility.
Finally, we compute the advantage $A(P)$ for each summary by normalizing its cumulative reward $R(P)$ against the mean and standard deviation of all rewards from the same rollout set.

\paragraph{Policy Optimization.}
Finally, the policy $\pi_{\theta}$ is then updated by optimizing the following clipped surrogate objective:
\begin{equation} \label{eq:ppo_loss}
\begin{aligned}
    \mathcal{L} = & -\frac{1}{2G}\sum_{P} \frac{1}{L_{P}} \sum_{t=1}^{L_{P}}\min ( \frac{\pi_{\theta}|_t}{\pi_{\text{old}}|_t} A(P), \\
    & \quad \, \text{clip}(\frac{\pi_{\theta}|_t}{\pi_{\text{old}}|_t}, 1-\epsilon, 1+\epsilon) A(P) ),
\end{aligned}
\end{equation}
where $P\in\{P_{i}^{k_1}\}_{i=1}^G\cup\{P_{j}^{k_2}\}_{j=1}^G$ is a summary from the rollout set,  $A(P)$ is its corresponding advantage, and $L_{P}$ is the length of $P$ and its corresponding reasoning process, and ${\pi_{\theta}|_t}$ and ${\pi_{\text{old}}|_t}$ are the token-level probabilities from the current policy and the old policy for the $t$-th token, respectively.

\begin{table*}[!t]
  \centering
  \caption{Main results across nine benchmarks. (P-Soups are split into three preference dimensions: ``expertise'', ``informativeness'', and ``style''.) We compare models in three settings: direct sequence modeling, full-history preference inference, and streaming preference inference. All preference inference models (both full-history and streaming) use the Qwen3-8B$_{\text{non-thinking}}$ as the downstream model. Within each setting, \textbf{Bold} and \underline{Underline} mark the best and second-best results among models at the $\sim$8B scale. \grayit{Gray} score highlights models that are outperformed by the best-performing $\sim$8B model in the same column, including direct sequence models and larger preference inference models that do not show a performance advantage.} 
  \label{tab:main-result}
  \adjustbox{max width=\textwidth}{
  \begin{tabular}{lcccccccccc}
    \toprule
    \multirow{3}{*}{\textbf{Model}} & \multicolumn{3}{c}{\textbf{In-domain}} & \multicolumn{6}{c}{\textbf{Out-of-domain}}&\multirow{3}{*}{\textbf{Avg.}} \\
    \cmidrule(lr){2-4} \cmidrule(lr){5-10}
     & \textbf{MIND} & \textbf{Amazon} & \textbf{AlignX} &
    \textbf{MovieLens} & \textbf{PersonaMem} & \textbf{Info.} &
    \textbf{Style} & \textbf{Expertise} & \textbf{HiCUPID} &  \\
    &(Rec.)&(Rec.)&(R.S.)&(Rec.)&(R.S.)&(R.S.)&(R.S.)&(R.S.)&(R.G.)&\\
    \midrule
    \midrule
    \multicolumn{11}{c}{\textbf{Direct Full-history Sequence Models w/o Preference Inference}} \\
    Qwen3-8B$_{\text{non-thinking}}$ & \grayit{63.03} & \grayit{84.05} & \grayit{59.63} & 88.57 & 61.40 & \grayit{46.84} & \grayit{42.33} & \grayit{38.33} & \grayit{47.02} & \grayit{59.02} \\
    TALLRec      & 81.96 & 94.91 & \grayit{66.30} & 97.90 & 64.36 & \grayit{51.66} & \grayit{70.16} & \grayit{60.16} & \grayit{47.41} & \grayit{70.53} \\
    \midrule
    \midrule
    \multicolumn{11}{c}{\textbf{Full-history Preference Inference}} \\
    DeepSeek-R1-671B         & \grayit{65.53} & \grayit{82.15} & \grayit{65.90} & 82.76 & 61.44 & \grayit{72.59} & \grayit{85.66} & \grayit{82.33} & 63.90 & \grayit{73.58} \\
    Qwen3-32B$_\text{thinking}$       & \grayit{67.63} & \grayit{85.69} & \grayit{64.93} & \grayit{75.43} & \grayit{57.36} & \grayit{73.25} & 88.00 & 83.66 & 63.44 & \grayit{73.26} \\
    GPT-OSS-20B              & \grayit{64.16} & \grayit{83.75} & \grayit{55.63} & \grayit{74.46} & 61.74 & \grayit{68.77} & \grayit{86.00} & \grayit{81.66} & \grayit{62.00} & \grayit{70.90} \\
    \midrule
    Qwen3-8B$_\text{thinking}$        & \underline{66.10} & \underline{84.68} & 62.73 & \underline{75.13} & \underline{54.36} & 75.08 & \textbf{87.50} & \textbf{83.50} & \underline{60.05} & \underline{72.12} \\
    DS-R1-Distill-Qwen-7B    & 61.20 & 82.82 & 54.03 & 70.30 & 49.28 & 56.14 & 65.83 & 66.00 & 60.01 & 62.84 \\
    \textsc{AlignXplore}              & 61.23 & 78.58 & \underline{66.60} & 69.93 & 53.98 & \underline{76.24} & 78.00 & 72.66 & 53.50 & 66.07 \\
    \textsc{AlignXplore+}                     & \textbf{71.36} & \textbf{86.39} & \textbf{75.03} & \textbf{75.80} & \textbf{58.08} & \textbf{78.07} & \underline{86.33} & \underline{82.50} & \textbf{62.42} & \textbf{75.10} \\
    \midrule
    \midrule
    \multicolumn{11}{c}{\textbf{Streaming Preference Inference}} \\
    DeepSeek-R1-671B         & \grayit{64.30} & \grayit{80.54} & \grayit{64.06} & 83.63 & 58.96 & \grayit{66.61} & 85.00 & \grayit{79.00} & \grayit{60.32} & \grayit{72.93} \\
    Qwen3-32B$_\text{thinking}$       & \grayit{66.60} & 85.35 & \grayit{64.60} & 77.78 & \grayit{53.26} & \grayit{73.58} & \grayit{83.66} & \grayit{81.67} & 59.83 & \grayit{71.81} \\
    GPT-OSS-20B              & \grayit{64.93} & \grayit{84.55} & \grayit{56.86} & \grayit{73.66} & 54.82 & \grayit{69.93} & \grayit{83.00} & \grayit{77.50} & \grayit{59.93} & \grayit{69.46} \\
    \midrule
    Qwen3-8B$_\text{thinking}$        & \underline{66.13} & \underline{83.58} & 62.90 & \underline{75.97} & \underline{51.68} & 74.08 & \textbf{85.00} & \textbf{82.66} & 59.17 & \underline{71.24} \\
    DS-R1-Distill-Qwen-7B    & 61.16 & 81.78 & 56.40 & 69.63 & 46.64 & 58.63 & 60.83 & 64.16 & \underline{59.29} & 62.05 \\
    \textsc{AlignXplore}              & 60.66 & 79.01 & \underline{69.90} & 67.96 & 48.42 & \underline{74.41} & 74.83 & 69.16 & 50.34 & 66.07 \\
    \textsc{AlignXplore+}                     & \textbf{71.80} & \textbf{85.35} & \textbf{73.67} & \textbf{77.23} & \textbf{54.58} & \textbf{76.57} & \underline{80.33} & \underline{78.50} & \textbf{60.51} & \textbf{73.17} \\
    \bottomrule
  \end{tabular}}
\end{table*}

\section{Experiments}

\subsection{Experimental setup}

\paragraph{Training Data.}
For SFT, we synthesize 540K training instances based on three datasets including Amazon-Book~\citep{amazon2018dataset}, MIND~\citep{mind2020dataset} and AlignX~\citep{alignx}, which span a wide spectrum of domains (e.g., e-commerce, news, daily life) and preference complexities.
During RL, we sampled 112K instances from the original datasets given limited computation resources. 
We provide more data details in Appendix~\ref{app:rl_data_implementation}. 

\paragraph{Implementation Details.}
We use Qwen3-8B~\cite{qwen3technicalreport} as the backbone. For SFT, we use PyTorch and the HuggingFace trl library with bfloat16 precision, Flash-attention-2 and TF32 arithmetic. 
For RL training, we use the VERL library. During evaluation, the default downstream model is Qwen3-8B. More implementation details are presented in Appendix~\ref{app:implementation}.

\paragraph{Benchmarks and Metrics.}
We evaluate our method on nine benchmarks spanning three distinct personalization tasks, evaluated by accuracy (ACC) unless stated otherwise: \textbf{(1) Recommendation (Rec.):} Predicting a user's preferred item from a pair based on their browsing or click history. Benchmarks include MIND~\cite{mind2020dataset}, Amazon-Book~\cite{amazon2018dataset}, and MovieLens-32M~\citep{movielens32m}.
\textbf{(2) Response Selection (R.S.):} Selecting the user-preferred response from candidates. This is tested on datasets with varied history formats, from structured preference triplets (AlignX~\cite{alignx}, P-Soups~\cite{psoups}) to raw dialogues (PersonaMem~\cite{personamem}).
\textbf{(3) Response Generation (R.G.):} Generating a personalized response from a conversational log, simulating an AI assistant, evaluated on HiCUPID~\citep{hicupid} using its benchmark-specific automatic evaluation.

\paragraph{Baselines.}
We compare our method against two categories of approaches:
\textbf{(1) Direct Sequence Models w/o Preference Inference:} This category includes models that process the raw user interaction history directly to make predictions, without an explicit preference summarization step. We employ {Qwen3-8B$_{\text{non-thinking}}$} and {TALLRec}~\citep{tallrec}, a strong baseline fine-tuned from Qwen3-8B on our SFT data. \textbf{(2) Preference Inference Models:} This category comprises LLMs that are prompted to first generate a textual preference summary, which is then used for the downstream task. These models span different scales, including 8B-sized models (AlignXplore~\citep{alignxplore}, Qwen3-8B$_{\text{thinking}}$, DeepSeek-R1-Distill-Qwen-7B~\citep{deepseekr1}), and state-of-the-art larger models ({Qwen3-32B}, {GPT-OSS-20B}~\citep{gptoss}, {DeepSeek-R1-671B}). Our model and all baselines are evaluated in both the \textbf{full-history} setting (processing the entire history at once) and the \textbf{streaming} setting (processing history in sequential segments). We provide further analyses in Appendix~\ref{app:further_analysis}.

\subsection{Main Results}
Our main results, presented in Table~\ref{tab:main-result}, lead to three key findings: \textbf{(1) Our framework outperforms similar-sized baselines in average score, and often exceeds even significantly larger models.} Even though our preference inference approach deliberately forgoes the full raw history available to direct sequence models, relying instead on a compact textual summary, it still consistently outperforms them. This demonstrates that the quality of representation matters more than the quantity of raw data, validating our framework's ability to distill user preferences into a highly efficient and comprehensive format.
\textbf{(2) Our framework's primary advantage lies in its capacity for deep preference reasoning, especially where user intent is subtle.} This is most evident on tasks like AlignX, where preferences must be inferred from conversational context. Our model achieves substantial gains over the next-best baselines (e.g., +8.43\% over AlignXplore in the full-history setting), while methods reliant on direct full-history sequence modeling struggle. 
We attribute this gain to our use of more comprehensive and heterogeneous training data, together with finer-grained reward signal, which enhances both the depth and coverage of the summarized preference. This demonstrates that by explicitly generating and validating textual hypotheses about user preferences, our model captures a deeper, more accurate understanding of user intent.
\textbf{(3) The explicit textual summary enables superior generalization across heterogeneous task formats, a critical challenge in personalization.} The contrast with TALLRec is particularly stark. While TALLRec is a powerful baseline for recommendation tasks, its performance collapses on the response generation task (HiCUPID), barely improving over a non-finetuned model. Our framework, however, maintains strong performance on HiCUPID, proving that a reasoned preference summary acts as a more robust and transferable representation of user interest than task-specific embeddings. This ability to bridge different tasks (e.g., from selection to generation) underscores the value of our text-based personalization approach.

\begin{table}[t]
  \centering
  \caption{Cross-domain transfer performance. We evaluate whether a preference summary generated from one domain can be applied to a different one under \textbf{full-history preference inference} setting. R.S.$\to$Rec.: summaries from AlignX (response selection) applied to MIND (recommendation), and vice versa.}
  \label{tab:transfer}
  \adjustbox{max width=\columnwidth}{
  \begin{tabular}{lcc}
    \toprule
    \textbf{Model} & \textbf{R.S.$\rightarrow$Rec.} & \textbf{Rec.$\rightarrow$R.S.} \\
    \midrule
    TALLRec  & 49.90 & 49.80 \\
    \midrule
    Qwen3-8B$_{\rm thinking}$ & 57.90 & 50.40 \\
    \textsc{AlignXplore+}     & \textbf{74.90} & \textbf{51.10} \\
    \bottomrule
  \end{tabular}}
\end{table}

\begin{table}[t]
  \centering
  \caption{Cross-model transferability of preference summaries. The table reports the average performance (\%) under \textbf{full-history preference inference} on recommendation (Rec.) and response selection (R.S.) tasks when summaries by different baselines are consumed by two distinct downstream models: Qwen2.5-7B-Instruct (top) and GPT-OSS-20B (bottom).  Full results are provided in Table~\ref{tab:qwen25} and \ref{tab:gptoss20b}.}
  \label{tab:reward-model-transfer}
  \adjustbox{max width=\columnwidth}{
  \begin{tabular}{lccm{0.01em}cc}
    \toprule
    \multirow{2}{*}{\textbf{Model}} & \multicolumn{2}{c}{\textbf{In-domain}} && \multicolumn{2}{c}{\textbf{Out-of-domain}} \\
    \cmidrule{2-3}
    \cmidrule{5-6}
    &\textbf{Rec.} & \textbf{R.S.} &&
    \textbf{Rec} & \textbf{R.S.} \\
    \midrule
    \multicolumn{6}{l}{\textbf{Downstream Model~=~Qwen2.5-7B-Instruct}} \\
    GPT-OSS-20B           & \grayit{73.39} & \grayit{55.86} && 75.23 & 69.42 \\
    DS-R1-671B      & \grayit{71.87} & \grayit{64.10} && 79.33 & 71.02 \\
    \midrule
    Qwen3-8B$_\text{thinking}$     & \underline{74.24} & 60.46 && \underline{73.20} & \underline{67.84} \\
    \textsc{AlignXplore}           & 69.33 & \underline{65.20} && 69.83 & 63.18 \\
    \textsc{AlignXplore+}       & \textbf{77.19} & \textbf{73.90} && \textbf{73.96} & \textbf{69.25} \\
    \midrule
    \midrule
    \multicolumn{6}{l}{\textbf{Downstream Model~=~GPT-OSS-20B}} \\
    GPT-OSS-20B           & \grayit{76.36} & \grayit{47.40} && \grayit{75.86} & \grayit{74.94} \\
    DS-R1-671B            & \grayit{70.05} & \grayit{68.60} && \grayit{77.33} & 78.65 \\
    \midrule
    Qwen3-8B$_\text{thinking}$     & \underline{76.58} & 52.23 && \underline{75.36} & \underline{73.61} \\
    \textsc{AlignXplore}           & 71.98 & \underline{64.73} && 69.13 & 70.97 \\
    \textsc{AlignXplore+}                  & \textbf{80.36} & \textbf{69.90} && \textbf{77.83} & \textbf{75.70} \\
    \bottomrule
  \end{tabular}}
\end{table}

\begin{table}[t]
  \centering
  \caption{User behavior transferability. The table reports the average performance (\%) on recommendation (Rec.) and response selection (R.S.) without negative samples in the history. Full results are provided in Table~\ref{tab:without-neg}.}
  \label{tab:without-negative}
  \adjustbox{max width=\columnwidth}{
  \begin{tabular}{lccccc}
    \toprule
    \multirow{2}{*}{\textbf{Model}} & \multicolumn{2}{c}{\textbf{In-domain}} && \multicolumn{2}{c}{\textbf{Out-of-domain}} \\
    \cmidrule{2-3}
    \cmidrule{5-6}
    &\textbf{Rec.} & \textbf{R.S.} &&
    \textbf{Rec.} & \textbf{R.S.} \\
    \midrule
    \midrule
    \multicolumn{6}{c}{\textbf{Direct Full-history Sequence Models}} \\ 
    Qwen3-8B$_\text{non-thinking}$      & \grayit{74.83} & \grayit{60.40} && 90.57 & \grayit{49.39} \\
    TALLRec                             & 88.50 & \grayit{60.46} && 97.30 & \grayit{58.34} \\
    \midrule
    \midrule
    \multicolumn{6}{c}{\textbf{Full-history Preference Inference}} \\
    GPT-OSS-20B                         & \grayit{76.61} & \grayit{55.23} && 85.33 & 69.22 \\
    Qwen3-32B$_\text{thinking}$         & \grayit{77.66} & \grayit{57.96} && \grayit{82.03} & 71.91 \\
    \midrule
    Qwen3-8B$_\text{thinking}$          & \underline{77.11} & 58.40 && \underline{80.66} & \underline{64.92} \\
    DS-R1-7B                            & 72.16 & 55.00 && 80.16 & 53.94 \\
    \textsc{AlignXplore}                         & 69.54 & \underline{61.40} && 75.40 & 63.19 \\
    \textsc{AlignXplore+}                                & \textbf{78.55} & \textbf{63.76} && \textbf{83.60} & \textbf{65.30} \\
    \bottomrule
  \end{tabular}}
\end{table}

\subsection{Transferable ability}
\label{sec:transfer}

A key motivation for our text-based personalization approach is transferability. Unlike task-specific latent representations, an explicit preference summary should function as a ``plug-and-play'' representation of user preferences. This allows it to be reused across different downstream models, applied to heterogeneous tasks, and remain robust even when the format of user interaction data varies. We evaluate our framework's performance across four critical dimensions of transferability.

\paragraph{Cross-domain Transfer.}
A truly unified user model should capture preferences that transcend the boundaries of a single application. We therefore test a critical scenario: can a preference summary, derived from a user's behavior in one domain (e.g., recommendation), be successfully applied to guide personalization for the {same user} in a completely different domain (e.g., dialogue)? This capability is vital for creating seamless, holistic user experiences across an ecosystem of services. 
As shown in Table~\ref{tab:transfer}, our method achieves substantially stronger cross-domain transferability than all baselines in both directions. The gains are especially pronounced when transferring rich dialogue preferences from AlignX to the MIND news recommendation task (see Appendix~\ref{app:cross_task} for dataset construction details). This suggests that our framework effectively decouples user's core preferences from the source domain, while the baselines lacks such a capability. 
The results also reveal an inherent asymmetry. Fine-grained conversational feedback provides richer and more transferable preference signals under shared topical spaces, whereas coarse, single-faceted news-click signals are inadequate for the more nuanced preference reasoning required in dialogue prediction. Consequently, although our model can infer wider and deeper preferences, the performance improvements from MIND (which offers relatively limited information) to AlignX (which demands richer preference signals) are necessarily modest.

\paragraph{Cross-model Transfer.}
We then evaluate if the generated summaries are useful for a variety of downstream models, not just the one used for training. 
Concretely, we first use the baselines and our \textsc{AlignXplore+} model to generate user preference summaries, and then feed these summaries into Qwen2.5-7B-Instruct and GPT-OSS-20B to perform downstream tasks.
Table~\ref{tab:reward-model-transfer} shows that while stronger downstream models naturally lead to better performance, our method consistently outperforms baselines across all tested models of a similar scale. This indicates that the quality of the summaries our framework produces is intrinsic and robust, providing a stable performance lift regardless of the specific downstream ``reader'' model. This is a crucial practical advantage over direct sequence methods like TALLRec, whose representations cannot be easily transferred to different model architectures.

\paragraph{User Behavior Transfer.}
We next evaluate the model's ability to transfer from its structured training format (paired feedback) to a more realistic positive-only format. This is crucial because explicit negative feedback is often unavailable in many real-world applications (e.g., browsing logs, liked items). To simulate this, we test the model's performance after removing all negative samples from the user histories. As seen in Table~\ref{tab:without-negative}, while all models experience a performance shift, our approach maintains its significant advantage over baselines, especially on response selection tasks where negative feedback is most informative. This successful transfer demonstrates that our framework's reasoning ability is not merely dependent on explicit negative signals, making it a robust and practical solution for real-world deployment.

\begin{figure}
    \centering
    \includegraphics[width=0.9\linewidth]{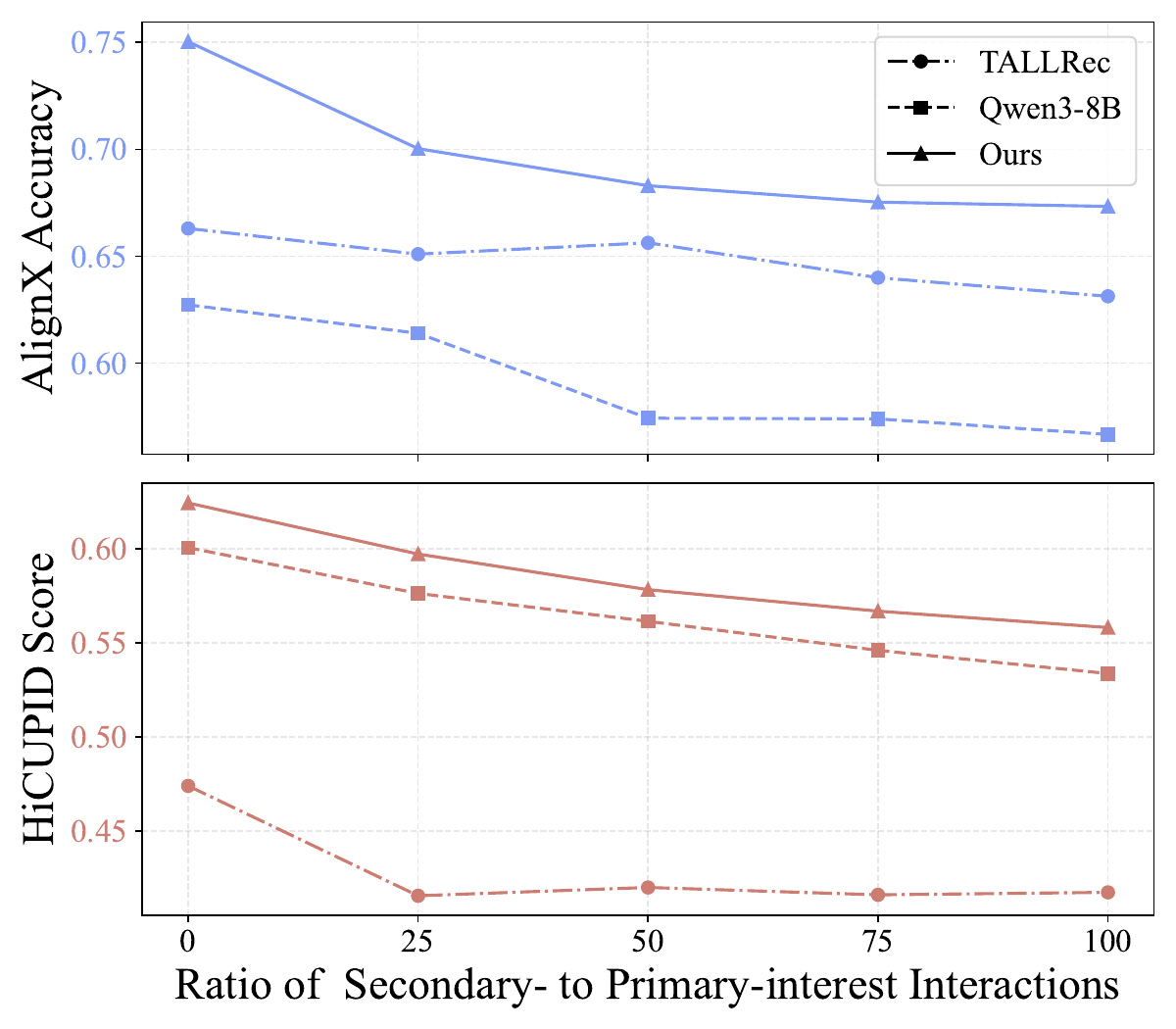}
    \caption{Multi-interest transferability. The figure plots the performance degradation on AlignX and HiCUPID with increasing history from a secondary interest. The x-axis represents the percentage of interactions from a secondary domain interleaved into the user's primary history.}
    \label{fig:noise}
\end{figure}

\paragraph{Multi-interest Transfer.}
A central challenge for a universal user model is its ability to synthesize a coherent profile from histories containing multiple, parallel interests. This scenario is critical, as modern users' digital footprints are fragmented across disparate services. To simulate this, we interleave a user's history from one domain with their history in another. Specifically, for response selection, we fuse user interactions from MovieLens into AlignX histories; for response generation, we similarly merge dialogue history concerning two different topics within HiCUPID (see Appendix~\ref{app:multi_interest} for dataset construction details). As illustrated in Fig.~\ref{fig:noise}, while this fusion of interests challenges all models, our method consistently maintains a significant performance advantage. This indicates our textual summary can effectively disentangle and represent distinct interests simultaneously, rather than conflating them into a noisy average. This capability is crucial for building a truly holistic user model suitable for cross-application personalization.

\subsection{Ablation study}
\label{sec:albation}

\begin{table}[t]
  \centering
  \caption{Ablation results of average performance (\%) of different variants on in-domain / out-of-domain recommendation (Rec.) and response selection (R.S.) tasks. Table~\ref{tab:ablation-full} in Appendix~\ref{app:full_results} shows the detailed results.}
  \label{tab:ablation}
  \adjustbox{max width=\columnwidth}{
  \begin{tabular}{lccccc}
    \toprule
    \multirow{2}{*}{\textbf{Model}} & \multicolumn{2}{c}{\textbf{In-domain}} && \multicolumn{2}{c}{\textbf{Out-of-domain}} \\
    \cmidrule{2-3}
    \cmidrule{5-6}
    &\textbf{Rec.} & \textbf{R.S.} &&
    \textbf{Rec.} & \textbf{R.S.} \\
    \midrule
    \midrule
    \multicolumn{6}{c}{\textbf{Full-history Preference Inference}} \\
    w/o RL                  & 76.87 & 65.03 && \underline{74.87} & 74.07 \\
    w/o SFT                 & 76.57 & 64.66 && 74.66 & \underline{75.35} \\
    w/o $R_{f}$    & \textbf{78.95} & \underline{73.66} && 74.53 & 75.05 \\
    \textsc{AlignXplore+}                    & \underline{78.88} & \textbf{75.03} && \textbf{75.80} & \textbf{76.25} \\
    \midrule
    \midrule
    \multicolumn{6}{c}{\textbf{Streaming Preference Inference}} \\
    w/o RL                  & 76.30 & 62.50 && 76.40 & 69.21 \\
    w/o SFT                 & 75.82 & 66.46 && 76.93 & 71.39 \\
    w/o $R_{f}$             & 78.11 & 71.03 && 75.43 & 71.53 \\
    \textsc{AlignXplore+}                    & \textbf{78.58} & \textbf{73.67} && \textbf{77.23} & \textbf{72.50} \\
    \bottomrule
  \end{tabular}}
\end{table}

We conduct an ablation study to dissect the contribution of each key component in our framework involving the following variants: \textbf{(1) w/o RL}: This variant omits the RL stage and relies solely on the initial SFT. \textbf{(2) w/o SFT}: This variant bypasses the SFT stage, applying RL directly to the base language model. \textbf{(3) w/o $\boldsymbol{R_f}$}: This variant removes the future utility term ($R_f$) from our reward function, optimizing the RL stage for immediate reward only. Ablation for training data synthesis are elaborated in Appendix~\ref{app:data_ablation}.

From the results in Table~\ref{tab:ablation}, we derive the following key findings: \textbf{(1) SFT and RL play distinct, complementary roles in achieving optimal performance.}  Removing the RL stage (\textit{w/o RL}) causes a consistent and significant performance drop across all datasets, which confirms that RL is essential for refining the model's ability to reason about and extract subtle user preferences. Conversely, removing the SFT stage (\textit{w/o SFT}) is particularly detrimental on complex datasets like AlignX, where user intent is nuanced. This suggests that SFT provides a critical ``cold start,'' equipping the model with a foundational understanding of preference structures, which RL then builds upon to achieve a deeper level of reasoning.
\textbf{(2) Modeling future utility ($\boldsymbol{R_f}$) is vital for effective streaming summarization.}  The removal of $R_f$ leads to a more pronounced performance degradation in the streaming setting compared to the full-history setting (e.g., a larger drop on AlignX and Info.). This is because the cumulative reward incentivizes our model to generate concise and forward-looking summaries that are easier to update and reuse. Without this form of temporal regularization, the model becomes myopic, optimizing for the current step at the expense of long-term summary quality and reusability.

\section{Conclusion}

In this work, we pioneer a paradigm shift, championing text as a universal interface for user representation. Our \textsc{AlignXplore+} generates textual summaries that not only achieve state-of-the-art performance, but more critically, exhibit unprecedented transferability across diverse tasks, models, interaction formats, and complex multi-interest histories. By transforming user profiles from opaque vectors into interpretable, portable narratives, our work paves the way for a new generation of transparent, interoperable, and truly user-centric AI agents.

\section*{Limitations}
While our framework demonstrates significant advantages in efficiency and effectiveness, several limitations warrant discussion to guide future research.
(1) Due to data limitations, we can only simulate transfer scenarios through synthesizing existing datasets, which may deviate from real-world deployment conditions to some extent. (2) Although we emulate a streaming setting, our method has not yet been deployed in a production environment. Therefore, its robustness under extremely long interaction histories remains unclear. (3) This work focuses on developing a relatively “universal” approach for modeling user preferences, and we therefore did not perform highly specialized or fine-grained design of the model training algorithms.

\section*{Ethics Statement}

All experimental datasets used in this study are derived from previously published work and obtained either via official APIs or by synthetic construction based on these sources. All user-related information in the datasets has been anonymized, and no personally identifiable information is involved. We do not use any non–open-source data. All data are used solely for scientific research, not for commercial purposes or for profiling or decision-making about individuals, and their acquisition and use comply with relevant ethical guidelines and standards of academic integrity. All existing resources used in our experiments—including datasets, pretrained models, and various APIs—are accessed and employed within the bounds of their original licenses and terms of use. The datasets and models we rely on have been filtered and processed by their original authors prior to public release to mitigate potential ethical risks.


\bibliography{clean}

\appendix

\section{Details for Supervised Fine-tuning Data Synthesis}
\label{app:sft_synthesis}
This section provides the technical details for our supervised fine-tuning data synthesis strategy, as introduced in Section~\ref{sec:cold-start}.

\subsection{History Segmentation and Target Selection}

For each user $u$ with an ordered interaction history $H = (h_1, h_2, \dots, h_K)$, we first partition it into non-overlapping, contiguous segments, $\{S_u^j\}$. For our experiments, we use three segments ($S_u^1, S_u^2, S_u^3$) corresponding to early, middle, and late periods of user activity. We filter out users for whom any segment contains fewer than a predefined number of interactions (e.g., 3).

The selection of ``target'' interactions is crucial. To identify meaningful targets, we first compute a \textit{Tractability Score}, $S_{\text{tract}}(h_i)$, for each interaction $h_i$ (details in Section~\ref{sec:data-pruning}). 
This score reflects how well a strong model can predict the user's choice. We extract a tractable subset $\bar{S}_u^j = \{h_i \in S_u^j | S_{\text{tract}}(h_i) \ge \tau_{\text{tract}}\}$, where $\tau_{\text{tract}}$ is a threshold. From each $\bar{S}_u^j$, we randomly sample a small number of interactions to form the target set $T_u^j$. These targets serve as anchors for profile generation and as ground-truth labels for our self-correction step.

\subsection{Pipeline Details}

The process can be formalized as follows:
\begin{enumerate}
    \item \textbf{Target-aware Generation}: For a history segment $S_u$ and its corresponding target set $T_u$, we generate a profile candidate $p_{u,t}$ for each target $t \in T_u$. If a prior profile $p_u^{\text{prev}}$ exists (from a previous segment), it is included in the context. The generation is performed by a teacher LLM, denoted by $\mathcal{G}$:
    \begin{equation}
        r_{u,t}, p_{u,t} = \mathcal{G}(\text{prompt}(S_u, t, p_u^{\text{prev}}))
    \end{equation}
    where $r_{u,t}$ is the chain-of-thought reasoning and $p_{u,t}$ is the resulting profile summary.

    \item \textbf{Preference-level Filtering}: For each pair $(p_{u,t}, t)$, we use a judgment model $\mathcal{J}$ to predict the user's choice in $t=(p_{\text{item}}, n_{\text{item}}, c)$. The prediction $\hat{y}_{u,t}$ is compared to the ground-truth choice $y_{u,t}$.
    \begin{equation}
        \hat{y}_{u,t} = \mathcal{J}(\text{prompt}(p_{u,t}, t))
    \end{equation}
    We form a set of consistent profiles $\mathcal{P}_u = \{p_{u,t} | \hat{y}_{u,t} = y_{u,t}, \forall t \in T_u \}$. If the size of $\mathcal{P}_u$ is smaller than a predefined threshold, we discard this user and proceed to the next user from Step 1.

    \item \textbf{Merging}: The consistent profiles in $\mathcal{P}_u$ (and their corresponding reasonings $\mathcal{R}_u$) are merged by a separate merging function, also implemented via a teacher LLM $\mathcal{G}_{\text{merge}}$:
    \begin{equation}
        p_u^{\text{merged}} = \mathcal{G}_{\text{merge}}(\text{prompt}(\mathcal{P}_u, \mathcal{R}_u))
    \end{equation}

    \item \textbf{User-level Filtering}: We compute an accuracy score for the merged profile $p_u^{\text{merged}}$ by evaluating its predictive performance on a set of validation targets $\tilde{T}_u \subseteq T_u$.
    \begin{equation}
    \begin{aligned}
                \text{Acc}_u = &\frac{1}{|\tilde{T}_u|} \sum_{t \in \tilde{T}_u} \mathbb{I}(\\
                &\mathcal{J}(\text{prompt}(p_u^{\text{merged}}, t)) = y_{u,t})
    \end{aligned}
    \end{equation}
    We only retain the user's data for SFT if $\text{Acc}_u \ge \lambda$, where $\lambda$ is a quality threshold (e.g., 0.8).
\end{enumerate}

This rigorous, multi-stage process ensures that the synthesized profiles are accurate, comprehensive, and behaviorally grounded, providing a high-quality dataset for SFT.

\subsection{Detailed Implementation}

When selecting targets, we employ a Qwen3-1.7B~\footnote{\url{https://huggingface.co/Qwen/Qwen3-1.7B}} model fine-tuned on a direct prediction task to compute $S_{\text{tract}}$, and set the tractability threshold to $\tau_{\text{tract}} = 0.9$. We retain only users whose tractable subset size exceeds 3. For users whose tractable subset size is larger than 5, we randomly sample 5 interactions to obtain $T_u^j$.

We then adopt vanilla Qwen3-32B~\footnote{\url{https://huggingface.co/Qwen/Qwen3-32B}} as the generation model $\mathcal{G}$, the judgment model $\mathcal{J}$, and the merging model $\mathcal{G}_{\text{merge}}$ to run the four-step synthesis pipeline. At the preference-level filtering step, we discard users whose preference set satisfies $\lvert \mathcal{P}_u \rvert < 3$. At the user-level filtering step, we set the user-level filtering accuracy threshold to $\lambda = 0.8$. 

We conduct the pipeline twice to construct training data for streaming setting and finally synthesize 540K training samples based on three datasets including Amazon-Book~\citep{amazon2018dataset}, MIND~\citep{mind2020dataset} and AlignX~\citep{alignx}, which span a wide spectrum of domains (e.g., e-commerce, news, daily life) and preference complexities. The detailed dataset sizes are reported in Table~\ref{tab:dataset}.

\begin{table}[t]
  \centering
  \caption{Number of training samples for SFT and RL.}
  \label{tab:dataset}
  \scalebox{0.9}{
  \begin{tabular}{lcccc}
    \toprule
    & \textbf{MIND} & \textbf{Amazon} & \textbf{AlignX} & \textbf{\#Total} \\
    \midrule
    SFT & 204{,}884 & 326{,}574 & 9{,}258  & 540{,}716 \\
    RL  & 29{,}567  & 11{,}653  & 70{,}941 & 112{,}161 \\
    \bottomrule
  \end{tabular}}
\end{table}

\begin{table}[t]
  \centering
  \caption{Hyper-parameters for RL sampling on different datasets.}
  \label{tab:sample}
  \scalebox{0.95}{
  \begin{tabular}{lcc}
    \toprule
    Dataset & \textbf{$\alpha$} & \textbf{$S_{\text{tract}}$}\\
    \midrule
    Amazon & 0.4 & $[0.50,0.90]$ \\
    MIND   & 0.1 & $[0.99,1.00]$ \\
    AlignX & 0.1 & $[0.98,1.00]$ \\
    \bottomrule
  \end{tabular}}
\end{table}

\section{Details for RL Data Pruning}
\label{app:pruning_details}
This section provides the technical details for our curriculum-based data pruning strategy, as introduced in Section~\ref{sec:data-pruning}.

\subsection{Pipeline Details}

\paragraph{Model Setup.}
Our method employs two autoregressive language models: a strong model, $M_{\text{strong}}$, and a weak model, $M_{\text{weak}}$. The weak model ($M_{\text{weak}}$) is a general-purpose language model with no task-specific training. The strong model ($M_{\text{strong}}$) is created by fine-tuning $M_{\text{weak}}$ on the entire user interaction dataset $\mathcal{D}$. The fine-tuning objective is to minimize the negative log-likelihood of the ground-truth preferred item $p_m$ in a binary choice context $(c_m, p_m, n_m)$:
\begin{equation}
\begin{aligned}
    \mathcal{L}_{\text{strong}} = &-\mathbb{E}_{(H^{[0,m)}, h_m) \sim \mathcal{D}} \\
    &\left[ \log p_{M_{\text{strong}}}(p_m\succ n_m | H^{[0,m)}, c_m) \right].
\end{aligned}
\end{equation}

\paragraph{Score Definitions.}
For each potential training point $(H^{[0,m)}, h_m)$, we define two scores:
\begin{enumerate}
    \item \textbf{Tractability Score ($S_{\text{tract}}$):} This is the probability assigned by the strong model to the ground-truth preferred item. It measures how solvable the sample is for a knowledgeable model.
    \begin{equation}
        S_{\text{tract}} = p_{M_{\text{strong}}}(p_m\succ n_m|H^{[0,m)}_u, c_m).
    \end{equation}
    
    \item \textbf{Learning Potential Score ($S_{\text{learn}}$):} This is the log-ratio of the probabilities assigned by the strong and weak models. It quantifies the performance improvement gained through task-specific fine-tuning, indicating the sample's reliance on learned reasoning.
    \begin{equation}
        S_{\text{learn}} = \log \frac{p_{M_{\text{strong}}}(p_m\succ n_m|H^{[0,m)}_u, c_m)}{p_{M_{\text{weak}}}(p_m\succ n_m|H^{[0,m)}_u, c_m)}.
    \end{equation}
\end{enumerate}

\paragraph{Filtering and Selection.}
We select our reasoning-intensive dataset, $\mathcal{D}_{\text{reason}}$, by applying a three-step filtering process to all potential training points $(H^{[0,m)}, h_m)$:
\begin{enumerate}
    \item We first select those with a Learning Potential Score in the top $\alpha\%$ percentile.
    \item From the remaining samples, we filter for samples within the ``sweet spot'' of tractability by applying both a lower and an upper bound on the Tractability Score: $\tau_{\text{low}} < S_{\text{tract}} < \tau_{\text{high}}$.
\end{enumerate}
For each user history $H$ containing two or more such filtered points, we rank these points in ascending order of their Tractability Score (i.e., from hardest to easiest). We then select the indices of the two hardest tractable points as our $k$ and $K$ to construct the final RL training instance $(H^{[0,k)}, h_k, H^{[k,K)}, h_K)$. 

\subsection{Detailed Implementation}
\label{app:rl_data_implementation}
During data pruning, we select ``easy'' examples from the AlignX and MIND datasets, while for the Amazon dataset we sample comparatively more challenging examples. We use Qwen3-1.7B and its fine-tuned variant as the ``weak'' and ``strong'' models to calculate scores. At filtering step, we first select the top $\alpha\%$ of examples according to $S_{\text{learn}}$, then apply an additional interval-based filter on $S_{\text{tract}}$, and finally take the top (easiest) or bottom (hardest) $c\%$ of the remaining examples according to $S_{\text{tract}}$. The specific sampling configurations are summarized in Table~\ref{tab:sample}. Under a fixed training budget, we proportionally sample approximately 112K instances from the original datasets according to their relative sizes. The detailed dataset statistics are reported in Table~\ref{tab:dataset}.

\section{Implementation Details}
\label{app:implementation}

We adopt Qwen3-8B~\footnote{\url{https://huggingface.co/Qwen/Qwen3-8B}} as the backbone. For SFT, we use PyTorch~\footnote{\url{https://pytorch.org/}} and the HuggingFace \texttt{trl}~\footnote{\url{https://huggingface.co/docs/trl/}} library with bfloat16 precision, Flash-attention-2~\footnote{\url{https://github.com/Dao-AILab/flash-attention}} and TF32 arithmetic. The backbone is fine-tuned for 4 epochs with a batch size of 16 and a maximum sequence length of 16,384 tokens. We employ the AdamW optimizer with a cosine learning rate schedule, setting the peak learning rate to $2 \times 10^{-5}$ and a warmup ratio of $0.1$. 
For RL training, we use the VERL~\footnote{\url{https://github.com/volcengine/verl}} framework, and set the rollout number to 4, the rollout batch size to 128, the training batch size to 512, the epoch number to 1, and the policy learning rate to 1e-6. We employ left truncation with maximum prompt and response lengths of 8192 and 4096, respectively. The policy is initialized from the SFT checkpoint, with gradient checkpointing. Training is distributed over 8 nodes with 8 GPUs each for one epoch as 109 steps.

For \textbf{full-history preference inference}, we directly feed the model with the user's complete interaction history to obtain a single preference summary. For \textbf{streaming preference inference}, we split each user's history into two halves: we first infer preliminary preferences from the first half, and then use this summary as initial preference when processing the second half of the history to derive the final preference summary. Finally, we condition the downstream models on both the inferred user preference and the task query to predict the correct answer. During evaluation, the default downstream model is Qwen3-8B.

\section{Transfer Data Synthesis}

This section explains the benchmark synthesis details for the transfer experiments, as introduced in Section~\ref{sec:transfer}.

\subsection{Cross-domain Transfer}
\label{app:cross_task}

We construct cross-domain scenario to evaluate the task transferability of the generated preferences. Specifically, we build an augmented dataset by matching user histories between MIND and AlignX using dense embeddings model. Specifically, we use the Qwen3-Embedding-8B~\footnote{\url{https://huggingface.co/Qwen/Qwen3-Embedding-8B}} model, implemented via the HuggingFace \texttt{transformers} library, to encode user histories into dense vector representations. 

For each user, we first convert the history into a single string using a prompt formatting. We then feed this formatted string into Qwen3-Embedding-8B and obtain one embedding vector per user. This procedure is applied independently to all users in MIND and AlignX. After encoding, we obtain one embedding for each of the $A$ users in MIND and of the $B$ users in AlignX. We stack the user embeddings from MIND into a matrix of size $A \times d$ and those from AlignX into a matrix of size $B \times d$, where $d$ is the embedding dimension. We then compute pairwise similarities between MIND and AlignX users by multiplying the two matrices, yielding an $A \times B$ similarity matrix. Based on these similarity scores, we select the 1{,}000 most similar user pairs across the two datasets.

For each selected user pair (one from MIND and one from AlignX), we create two new testing examples by swapping their targets while keeping their own histories. From 1{,}000 similar user pairs, this procedure produces 2{,}000 new user-level instances to evaluate cross-domain transfer performance.

\subsection{Multi-interest Transfer}
\label{app:multi_interest}

Beyond evaluating the overall transferability of user-level preferences, we also examine the model’s generalization ability in interest-level transfer within a single user history. In realistic settings, a user may maintain multiple concurrent preferences across different platforms or domains. To approximate this setting, we construct synthetic multi-interest users by merging histories from datasets with low topical overlap while explicitly avoiding preference conflicts, and then assess whether the preference inference model can retain all parallel key preferences in the fused histories.

For the response selection task, we fuse user histories from MovieLens into AlignX. Specifically, for each user in AlignX, we associate a random counterpart user from MovieLens. For each such user pair, we randomly sample a subset of the MovieLens user's historical interactions, where the subset size is controlled by a ``noise intensity'' parameter. We define ``noise intensity'' as the proportion of sampled interactions that are irrelevant to the target among all interactions in that user's original history. We then insert the sampled noisy interactions into the original interaction sequence at random positions while preserving their relative order, resulting in noisy user histories for evaluation. For the response generation task, we first pair dialogues for different topics in HiCUPID dataset, and then apply the same procedure.

\section{Further Analyses}

This section provides further experimental results and analyses, including ablation for training data synthesis as mentioned in Section~\ref{sec:albation} and reward \& reponse length evolution.

\subsection{Data Ablation}
\label{app:data_ablation}

\begin{table}[t]
  \centering
  \caption{Average performance (\%) of models trained with 10\% sub-datasets constructed using our method and random sampling (RDM) on in-domain / out-of-domain recommendation (Rec.) and response selection (R.S.) tasks. Detailed results can be found in Table~\ref{tab:data_ablation}.}
  \label{tab:mini-random-sft-rl}
  \adjustbox{max width=\linewidth}{
  \begin{tabular}{lccccc}
    \toprule
    \multirow{2}{*}{\textbf{Model}} & \multicolumn{2}{c}{\textbf{In-domain}} && \multicolumn{2}{c}{\textbf{Out-of-domain}} \\
    \cmidrule{2-3}
    \cmidrule{5-6}
    &\textbf{Rec.} & \textbf{R.S.} &&
    \textbf{Rec.} & \textbf{R.S.} \\
    \midrule
    \midrule
    \multicolumn{6}{c}{\textbf{SFT-Only}} \\
    \midrule
    \multicolumn{6}{c}{\textbf{Full-history Preference Inference}} \\
    RDM  & 71.08 & 64.46 && 70.76 & 72.93 \\
    Ours & 71.01 & 65.20 && 70.36 & 74.06 \\
    \midrule
    \multicolumn{6}{c}{\textbf{Streaming Preference Inference}} \\
    RDM  & 70.74 & 65.13 && 71.13 & 69.69 \\
    Ours & 70.66 & 66.90 && 70.00 & 71.82 \\
    \midrule
    \midrule
    \multicolumn{6}{c}{\textbf{SFT \& RL}} \\
    \midrule
    \multicolumn{6}{c}{\textbf{Full-history Preference Inference}} \\
    RDM  & 70.61 & 65.43 && 69.00 & 73.77 \\
    Ours    & 71.68 & 67.83 && 69.23 & 76.09 \\
    \midrule
    \multicolumn{6}{c}{\textbf{Streaming Preference Inference}} \\ 
    RDM  & 70.23 & 65.67 && 69.73 & 70.26 \\
    Ours    & 70.69 & 68.40 && 70.03 & 73.11 \\
    \bottomrule
  \end{tabular}}
\end{table}

To validate the effectiveness of the data construction strategy in Section~\ref{sec:method}, we conduct an ablation study using only about 10\% of the full dataset for training in both the SFT and RL stages due to limited resources. Specifically, we use two approaches to conduct training dataset: (1) adopt our proposed data selection procedures (``divide-and-conquer'' SFT data synthesis and curriculum-based data pruning technique in RL); (2) perform random sampling without any filtering to obtain user instances as the baseline since it is common practice.

The results are reported in Table~\ref{tab:mini-random-sft-rl}, from which we have two main observations.
\textbf{(1) Our sampling strategy yields training data that helps the model capture wider and deeper user preferences.} This holds both when using SFT alone and when using the full two-stage training framework. Models trained on our selected data perform better on tasks with more complex and more implicit preference signals. On standard recommendation benchmarks, the gains are more moderate.
We attribute the improvement in the width of captured preferences mainly to the SFT data construction. It encourages the model to consider as many potential downstream “targets” as possible when generating preference summaries. The improvement in depth comes from the calibrated difficulty of the selected data. In the RL stage, our method avoids overly easy or hard “noisy” examples. This allows the model to explore diverse preferences within its capability and to receive more informative feedback.
\textbf{(2) In the SFT stage, our method produces data that provides a stronger foundation for subsequent RL.} When considering SFT alone, our approach does not show a clear advantage. However, after applying RL, the performance gains become substantially larger, especially on the response selection task. Stage~2 of our SFT data construction helps avoid incorrect preference summaries. As a result, it preserves the correctness of the model’s learned knowledge and supports more effective exploration during RL.

\begin{figure*}
    \centering

    \begin{subfigure}{0.45\linewidth}
        \centering
        \includegraphics[width=\linewidth]{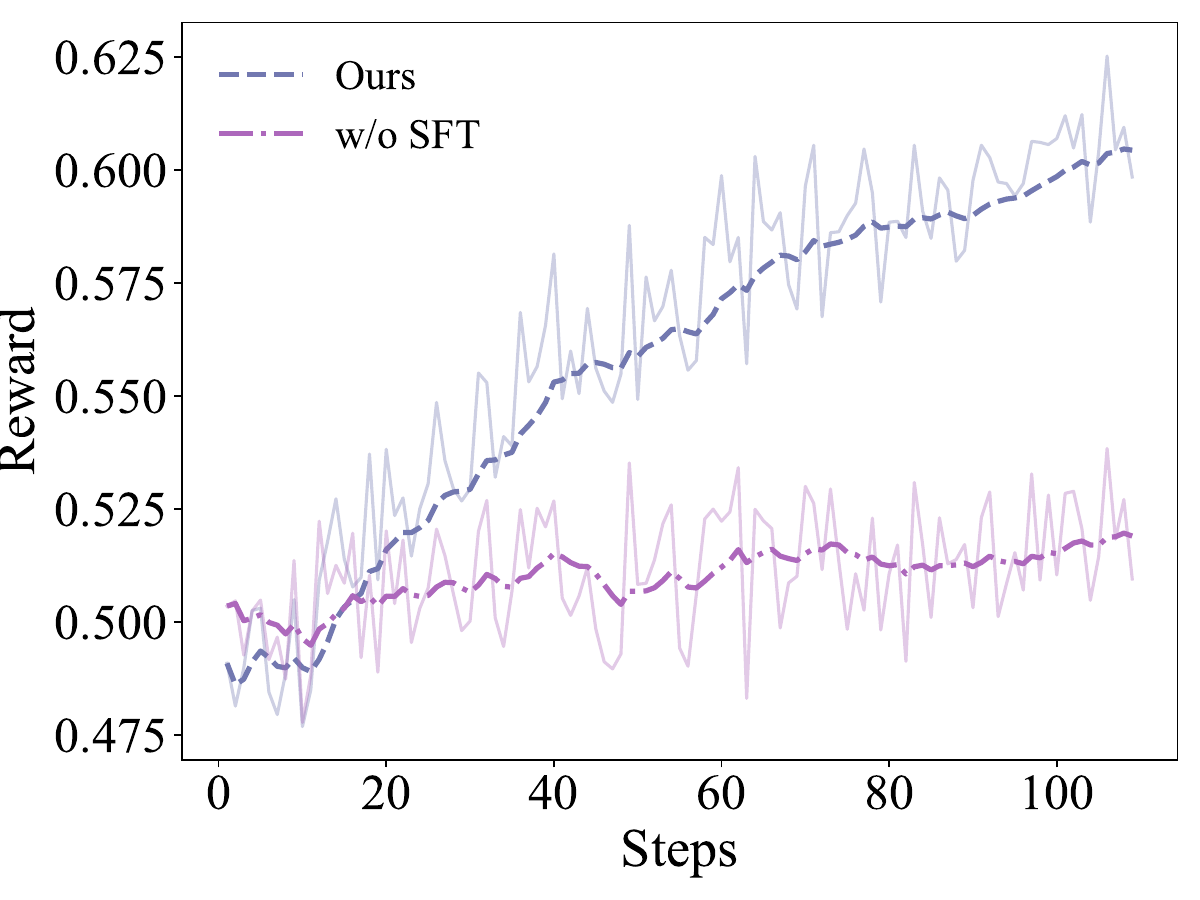}
        \label{fig:up}
    \end{subfigure}
    \begin{subfigure}{0.45\linewidth}
        \centering
        \includegraphics[width=\linewidth]{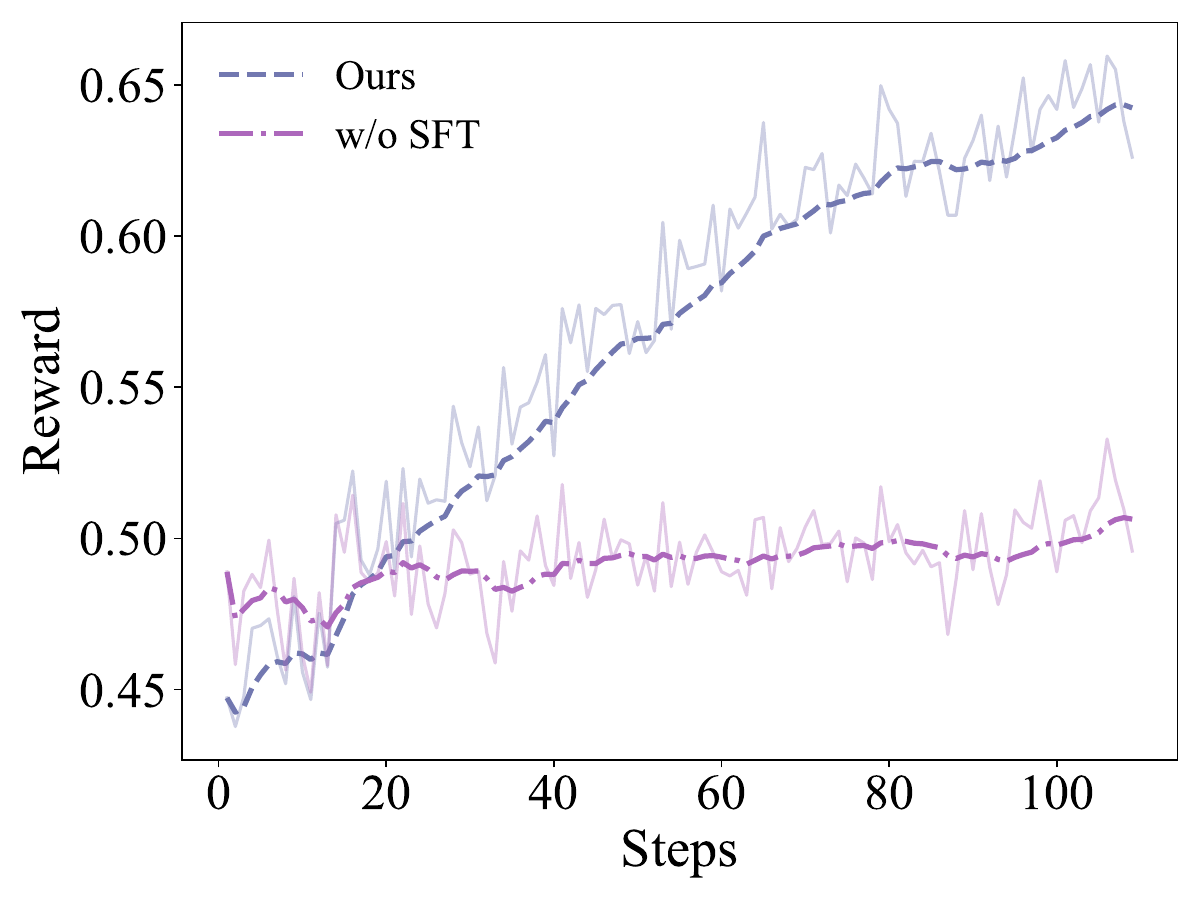}
        \label{fig:down1}
    \end{subfigure}
    \begin{subfigure}{0.45\linewidth}
        \centering
        \includegraphics[width=\linewidth]{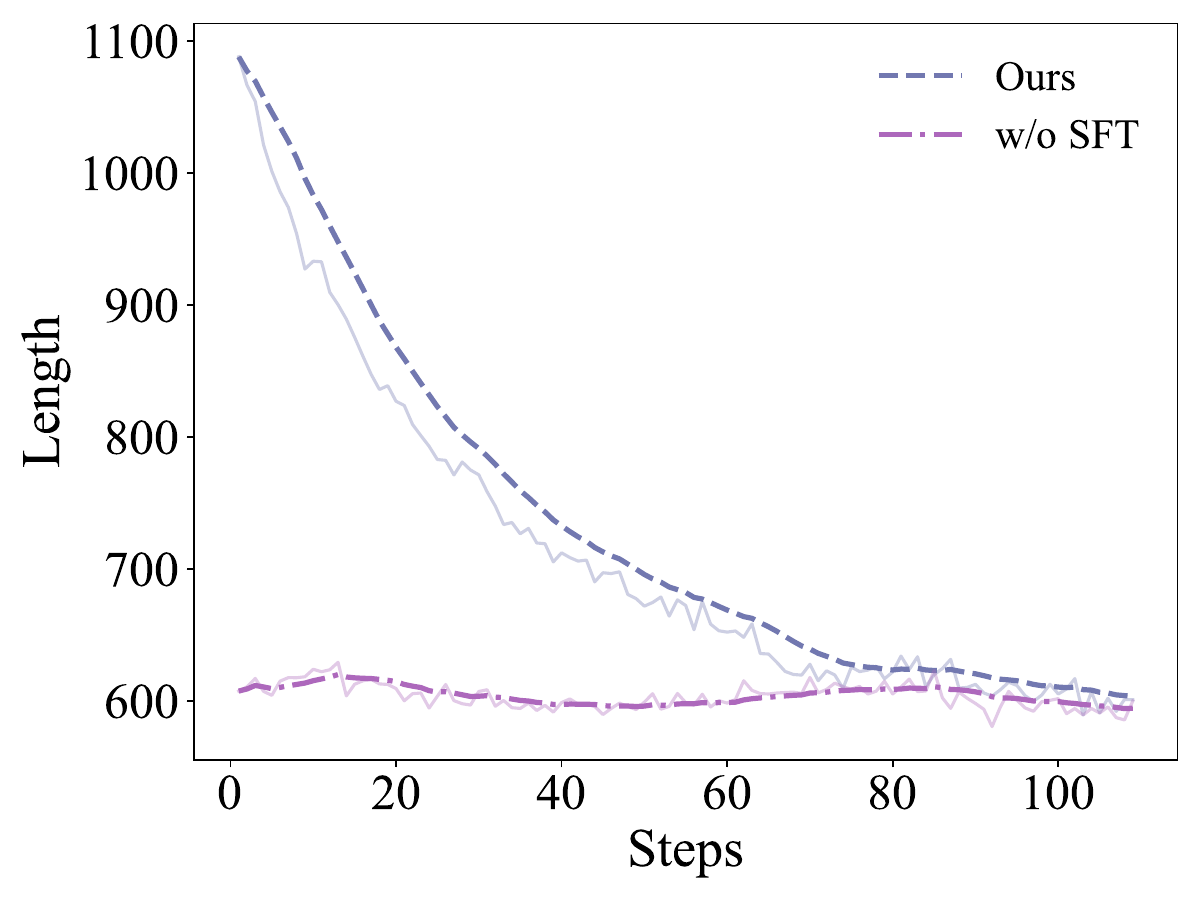}
        \label{fig:down2}
    \end{subfigure}
    \begin{subfigure}{0.45\linewidth}
        \centering
        \includegraphics[width=\linewidth]{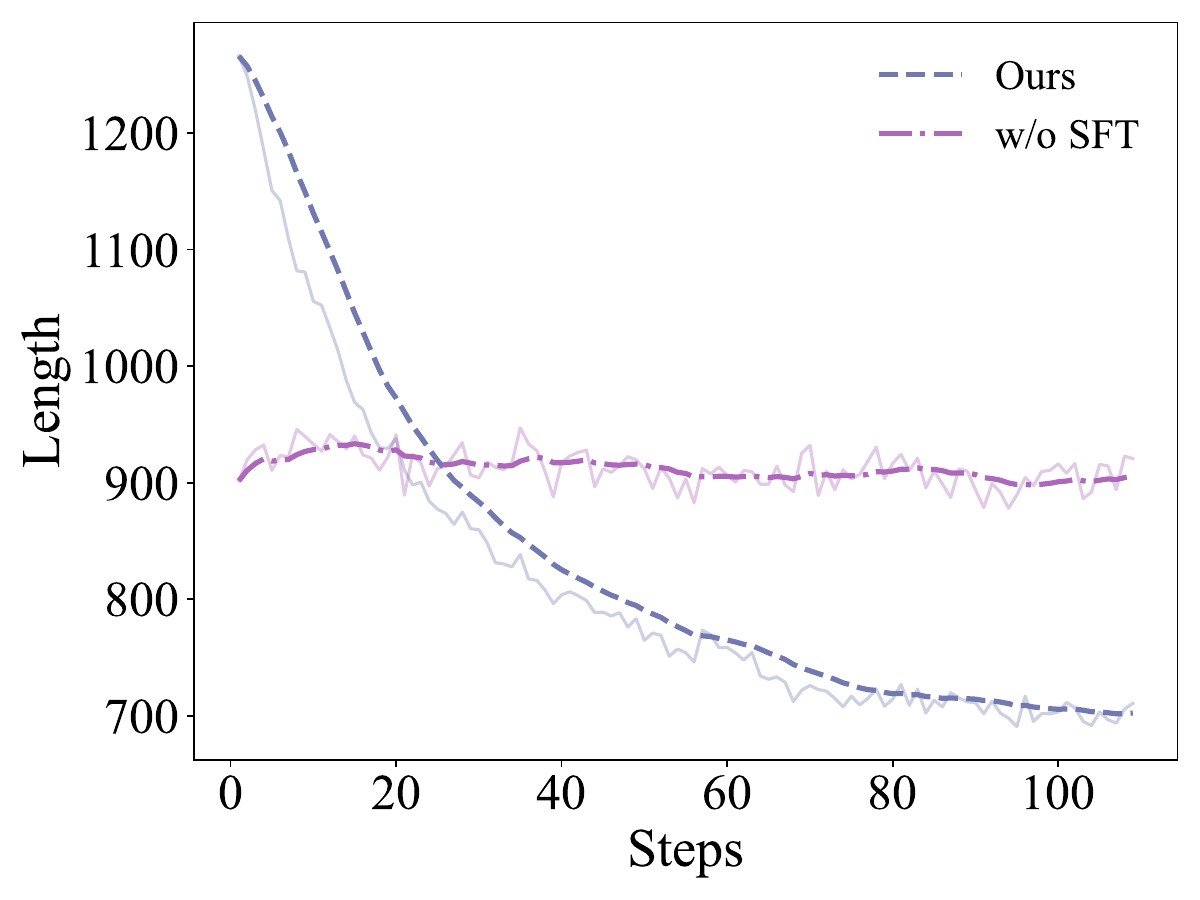}
        \label{fig:down3}
    \end{subfigure}
    \vspace{-20pt}
    \caption{RL training curves. Reward and response length of initial (left) and updated (right) preference summarizing.}
    \label{fig:reward-and-length}
\end{figure*}

\subsection{Reward and Response Length Evolution}
\label{app:further_analysis}

We further analyze how the reward and response length evolve with and without cold-starting. The results are illustrated in Figure~\ref{fig:reward-and-length}.

Without cold start, the rewards for both preference summary fail to increase steadily. Although the model attempts to explore, it cannot discover an effective optimization path under the RL objective because it has never been exposed to the training data format or distribution. The response length curves also reveal that, without SFT, the model tends to simply append a few additional preferences in the updating stage on top of the first-stage summary, instead of restructuring or refining the preference representation.

In contrast, the cold-started model can be effectively optimized under the RL setting, leading to stable reward growth in both stages. After SFT, the model’s responses become more comprehensive and longer at the beginning of training. During RL, the model gradually converges to preference expressions that are both concise and informative: the updated preferences evolve toward a length–quality trade-off that yields the highest reward, rather than naively accumulating more content. These observations indicate that our SFT + RL strategy is necessary to guide the model from initially broad, exhaustive preference descriptions toward compact yet comprehensive preference representations.

\section{Case Study}

We present a representative user case together with the inferred preferences generated by the baselines and by our \textsc{AlignXplore+} under different settings. We additionally highlight the key segments in the interaction history that are used for preference summarization. Under streaming inference with access to the full history, we make the following observations:

\begin{itemize}
    \item Several of the preferences generated by Qwen3-8B conflict with the underlying data or are only weakly supported. (1) It treats gun control as a core issue the user ``seeks,'' even though the long post on gun control in the history is labeled as rejected and thus should not be used to infer a central preference. (2) It claims that the user ``prioritizes experienced leadership over charisma or populism,'' whereas the selected examples explicitly argue that the ideal 2020 candidate should be ``charismatic, media savvy, relatively young'' and emphasize ``economic populism,'' directly contradicting this summary. (3) The assertion that the user prefers ``policy-driven candidates'' is also largely speculative. The evidence in the chosen items is much stronger on election viability and strategy, media narratives, rejection of false equivalence, and attitudes toward third parties and voting behavior than on systematic policy discussion per se.
    \item The preference summary generated by \textsc{AlignXplore} also contains several claims that are not well supported by the labeled data. (1) It states that the user selects content such as the ``GOP retreat / WP audio tapes,'' but this segment appears in the rejected set. The passage is long, includes a subscription call-to-action, and contains emotionally charged mockery (e.g., ``Cheeto…''), so it cannot be used to support the claim that the user prefers this type of content. (2) It overgeneralizes by asserting that the user rejects humor and emotional framing in favor of purely objective analysis. In the history, the user does select items with witty commentary (e.g., the ``February 23th / Twenty-thirth'' example), and the accepted items also contain clear emotional expressions such as ``pit in my stomach,'' ``makes me sick,'' ``go away already,'' and ``He is the worst.'' Thus, the claim that the user rejects emotional content and only values objectivity is not tenable. (3) It overstates the user’s supposed preference for policy or structural issues over symbolism. The more stable themes in the data concern election strategy and feasibility, critique of partisan narratives, rejection of false equivalence, and resistance to meme-ification, rather than sustained ``deep dives'' into policy. As a result, it inflates the degree of ``policy focus'' beyond what the evidence supports.
    \item The preferences inferred by our \textsc{AlignXplore+} exhibit several notable advantages. (1) They are more tightly grounded in the available evidence. The selected parts of the history are heavily concentrated on U.S. politics and current affairs, election probabilities and strategy, and critical attitudes toward political figures and governance (e.g., anxiety over FiveThirtyEight forecasts, repeated questioning of the Yemen raid, discomfort with ``both sides'' narratives, and frustration with third-party vote splitting). Our summary captures this central theme more accurately and consistently. (2) It makes fewer unsupported inferences about specific policy issues, avoiding the temptation to elevate one-off or rejected items into core preferences.
\end{itemize}

For full-history inference without access to negative samples, we obtain the following conclusions:

\begin{itemize}
    \item The preferences generated by Qwen3-8B remain too close to a generic ``politically engaged user'' template and underemphasize the user’s key ``red lines'' and rejection patterns. For example, it mentions ``balanced discourse'' and ``critical thinking,'' but the user’s strongest triggers are in fact opposition to ``both-sides'' false balance and to the normalization of extremism (e.g., Nazi violence). By failing to encode this as a firm preference, it may cause downstream assistants to mistakenly assume that the user appreciates ``neutral, balanced debate.'' The summary is also not sufficiently actionable: it describes what topics the user likes to discuss, but not which narratives the user finds off-putting or which styles of expression they respond well to. For instance, the history shows that the user clearly accepts sharp, sarcastic, and mocking critiques of Trump, yet this is not highlighted. Finally, the boundaries it encodes are blurry. By emphasizing ``media scrutiny'' and ``balanced discussion,'' it can inadvertently steer later outputs toward equivocation or ``both-sides-ism,'' whereas the history (especially in Trip 4) indicates that this is precisely the stance the user is most opposed to.
    \item The preferences generated by our \textsc{AlignXplore+} are more controllable and practically useful: they go beyond a thematic summary and can serve directly as an ``operational manual'' for downstream dialogue. (1) They specify what the assistant should provide: concrete data, causal pathways, probabilities, representative numbers, and analyses of real-world consequences. (2) They explicitly state what the assistant must avoid: false balance, ``both-sides-are-bad'' framings, downplaying extremism, and portraying third-party protest voting as a moral high ground. (3) They delineate acceptable tonal choices: sharp, sarcastic criticism of Trump is allowed—and in fact well received by the user—whereas enforcing an artificially ``objective'' stance is not required. Moreover, they offer broader coverage without losing focus. Whereas the preferences summarized by Qwen3-8B are presented in a more point-wise fashion, ours connect these points into a coherent set of rules. This rule-based structure makes our preferences more likely than the baselines’ to transfer robustly to new conversational scenarios.
\end{itemize}

Finally, we compare preferences inferred for the user’s secondary interests. Preference A (generated by our \textsc{AlignXplore+}) provides a more accurate characterization of the user’s movie preferences. It correctly captures the common pattern among multiple accepted items: mainstream, highly watchable action/thriller/comedy/sci-fi/adventure titles (e.g., \emph{In the Line of Fire}, \emph{Catch Me If You Can}, \emph{Independence Day}, \emph{Men in Black II}, \emph{Unbreakable}). The user repeatedly rejects documentaries, niche films, old animated shorts, and more ``offbeat'' entries (e.g., the 1939/1949 animations, the documentary \emph{Knife Skills}, \emph{If These Walls Could Sing}). By contrast, preference B (generated by Qwen3-8B) claims that the user ``avoids animation or children’s films,'' which is clearly incorrect: the user selects family-/children-oriented works such as \emph{Home Alone}, \emph{Charlie and the Chocolate Factory}, and \emph{Who Framed Roger Rabbit?}—the last of which is itself heavily driven by animation. On political content, A is also more accurate in capturing the user’s stylistic preferences. The user more often selects content with clear information and viewpoints, pragmatic framing, and grounding in electoral realities (e.g., anxiety over FiveThirtyEight probabilities, reminders to research nonpartisan options, questioning of the Yemen raid, and dissatisfaction with ``both sides'' narratives). In contrast, B incorrectly promotes gun control to a core issue: the segment on guns/Ramadan in the history is rejected, suggesting that strongly didactic, long emotional monologues of that sort are not to the user’s taste.

\onecolumn

\begin{tcolorbox}[colframe=gray!20!white, colback=gray!3!white, coltitle=black, fonttitle=\bfseries, title=User History in AlignX for Case Study, breakable]
1. \\
\textbf{Post}: You can use this thread to discuss whatever is on your mind or \grn{share news articles} or off topic things that would otherwise not be posted to the sub.
Morning Roundtable - 6/13

\textbf{Chosen}: \grn{The Supreme Court has declined to review a case on citizenship for residents of American Samoa.}

\grn{Not a great time to be a U.S. territory. Additionally, Puerto Rico intends to speak to the U.N. about the unfavorable decision from last week.}

\textbf{Rejected}: The real fight going forward for me is \grn{Gun Control.}  I am really tired of hearing that if you want to get elected, you need the NRA.  Why?  Is money more powerful than our votes?  Are they trying to say that our vote do not matter?  Is gun more important than human life?  In the month of Holy Ramadan, every Muslim should denounce terrorism.  And to my fellow Americans, please fight to stop the agony of losing loved one to gun violence.  One life lost to gun is too many.  Thanks

\ 
\\2. \\
\textbf{Post}: You can use this thread to discuss whatever is on your mind or share news articles or off topic things that would otherwise not be posted to the sub.
Roundtable - 7/12

\textbf{Chosen}: \grn{Has there been a single day without scandal during the Trump administration.}

\textbf{Rejected}: \grn{I'm watching Modern Family for the first time and I giggled with glee at a moment when a character says he donated some of his old clothes to somebody, and the next shot is the person wearing a ``HILLARY 200'' tshirt.}

\ 
\\3.\\
\textbf{Post}: You can use this thread to discuss whatever is on your mind or share news articles or off topic things that would otherwise not be posted to the sub.
Morning Roundtable - 9/10

\textbf{Chosen}: \grn{Trump's campaign is going to try to fight the Hillary camp on that comment but this is exactly the fight they want.}

\textbf{Rejected}: Everyone knows Trump's subreddit it a \grn{racist, misogynist Hate Group}, but who could guess they have such tender, delicate, easily hurt feelings? Just like Donald himself. He calls people names and spews hate constantly, but if anyone does not give him what he feels is his due deference for being a spoiled little rich boy all grown up, he has a wittle temper tantrum.

\ 
\\4.\\
\textbf{Post}: You can use this thread to discuss whatever is on your mind or share news articles or off topic things that would otherwise not be posted to the sub.
Roundtable - 8/14

\textbf{Chosen}: \grn{I find it disturbing how many people on Reddit are trying to handwave and dismiss actual murderous Nazis on American streets. There are literally people saying antifa caused the violence and totally ignoring the terrorist attack. People are fine with platitudes but when you come out and flatly say that the Nazis are a problem they start wringing their hands that ``both sides are bad!''}

\textbf{Rejected}: \grn{``Nazis are lot like Cats, if they like you is probably because you're feeding them''}

-John Oliver

\ 
\\5.\\
\textbf{Post}: I just wanted to say hello and tell all of you that I am new to this sub and I couldn't be happier.  I can no longer handle the insanity that has taken over Reddit.  If I see one more delusional Sanders article I'm going to jump off a bridge.  I've been a long time supporter of HRC and I have felt isolated and alone on this site for sometime.  Thank you to whoever made this sub happen.
Thank you for being here.

\textbf{Chosen}: Hey guys - I'll preface this post by saying I'm an Australian so I'm an outsider to the US elections, and my main source of information about the presidential race is Reddit.

Thus, I hear a lot of good stuff about Bernie, some good and some bad about Trump, and a LOT of negativity about Hillary. If /r/sandersforpresident were to be believed, one would wonder how Hillary is even a serious contender at this point with all the shit that she's taking flak for. So, can you guys please \grn{fill me in}?

\grn{Is this website just incredibly biased and misinforming me about Hillary? What are her main stances and why is she still in the race when she is basically portrayed as the anti-christ?}

I appreciate any helpful info or discussion because obviously Hillary mustn't be as bad as I've seen if she's still in the lead (I think?).

\textbf{Rejected}: \grn{I'm just worried that HC won't be able to compete with the memes being pumped out by TD and S4P.  I think we can all agree that memes are the way a part of the youth vote declares their alliances.  Viral memes will make a candidate go viral.}

This sub needs to embrace the insanity and steer it in the right direction.

\ 
\\6.\\
\textbf{Post}: You can use this thread to discuss whatever is on your mind or share news articles or off topic things that would otherwise not be posted to the sub.
Afternoon Roundtable - 9/26

\textbf{Chosen}: \grn{538 ``Now-cast'' shows Hillary below 50\% probability. This past week is the first time I actually have a pit in my stomach about this election. I am also very worried about my fellow millennials going to third parties. The idea of a liberal in Florida voting for Jill Stein makes me sick, honestly.}

\textbf{Rejected}: \grn{Drinking game ideas for tonight?}

Take a drink every time

* Bill Clinton's infidelity is discussed

* Trump pretends to care about Bernie Sanders

* Benghazi is mentioned

* Emails are mentioned

* Trump blames Hillary for something he did

* Trump unironically calls the Clinton Foundation a ``criminal enterprise''

Drink the whole bottle if Trump doesn't actually show up.

\ 
\\7.\\
\textbf{Post}: You can use this thread to discuss whatever is on your mind or share news articles or off topic things that would otherwise not be posted to the sub.
Evening Roundtable - 8/17

\textbf{Chosen}: Watching Rachel Maddow, they showed a screenshot of \grn{what the Trump website's Nevada page used to look like}. It told you to go and vote [in the primary] on \grn{February 23th}.

February 23th

Twenty-thirth. Y'know, the famous date of February twenty-thirth.

\textbf{Rejected}: \grn{Jill Stein is like the mom who wants to be cool , so she starts trying to act young and hip.}

\ 
\\8.\\
\textbf{Post}: You can use this thread to discuss whatever is on your mind or share news articles or off topic things that would otherwise not be posted to the sub.
Morning Roundtable - 9/03

\textbf{Chosen}: My favorite part of Trump's speech at the church:

\grn{``I wrote this a few days ago when I found out I would be speaking here... And I think you'll understand it, maybe, even better than I do in certain ways.''}

Proceeds to read something he obviously didn't write himself.

\textbf{Rejected}: I want to play a fun game with you guys. If Clinton gets elected how many investigations do you thing a Republican Congress holds within the first year? Over or Under 50?

\ 
\\9.\\
\textbf{Post}: You can use this thread to discuss whatever is on your mind or share news articles or off topic things that would otherwise not be posted to the sub.
Roundtable - 11/20

\textbf{Chosen}: \grn{Thoughts on Tim Ryan? Does he make his case against Nancy Pelosi?}

\textbf{Rejected}: \grn{after all this news about how shitty men are (i myself am a man): i propose the following to fix our country:}

\grn{no more male suffrage}

i'm joking, but seriously i think for the good of the world, it's time for men to step back and for women to take charge.

\ 
\\10.\\
\textbf{Post}: You can use this thread to discuss whatever is on your mind or share news articles or off topic things that would otherwise not be posted to the sub.
Roundtable - 2/02

\textbf{Chosen}: \grn{Where is the outrage about the raid in Yemen that got a Navy SEAL and an eight year old girl killed?}

\textbf{Rejected}: \grn{Trump is like the main villain from a crappy gritty reboot of a cartoon.}

\ 
\\11.\\
\textbf{Post}: You can use this thread to discuss whatever is on your mind or share news articles or off topic things that would otherwise not be posted to the sub.
Roundtable - 1/29

\textbf{Chosen}: \grn{I have a feeling HIllary's gonna run again in 2020, which would be welcome news to me.} She deserves the chance to see her dream come to fruition, espescially after the witch hunt she was the subject of  this cycle. Everyone says ``oh no I don't want to see her go through that again'' or ``we need a new face'' but I say a) she's strong enough to handle it b) she deserves to at least try and c) we don't need a fresh face, we need \grn{someone who knows how to get the job done}, and I believe that person is \grn{Hillary}

\textbf{Rejected}: I tried to post the link to it but it was auto removed by the mods. The WP has posted all the GOP audio tapes online...

The Post and Times are holding the line and Trump is triggered, again. He is tweeting about them losing subscribers (not true). Instead of acknowledging the first soldier to die on his watch and the nation protesting, he tweets for his ego.

Next he will cry about the WP again.

WP has made the full tapes but I can not post the link. It is the verified GOP retreat unfiltered and unabridged. It is clear that all the phone calls, visits and letters are working. If you have used all your WP views this month, search for it in google news and use that link.

Consider subscribing to the NYT and WP to support them. Cheeto, our Presidents Bannon and Miller and their smiling clown Pence can't stand them not rolling over!

If you already have an Audible subscription you have a credit for a free daily audio version of either the NYT or the WP. You must select which paper you prefer in your membership screen. You can also switch them at any time. Once turned on, the audio is auto delivered to your device or app each morning. You can catch up on your commute and give Trump a finger at the same time, no extra charge. (The NYT has the better narrator in my opinion)

\ 
\\12.\\
\textbf{Post}: Wisconsin Sanders +6
New York Clinton +12

Sanders new targets require him to win both. Wisconsin by 16 and New York by 4. These are two of the largest states left in delegates, and WI is the largest Sanders favorable state left. Him winning it by 6 is a nightmare for him, as is losing NY by any margin, especially double digits as shown here.

If these results don't improve dramatically in 3 weeks, 350 more delegates will be removed from the board and Hillary will take the majority of them yet again. His path will narrow from unlikely to extremely unlikely, perhaps impossible.
New Polls Today Discussion: Wisconsin \& New York

\textbf{Chosen}: \grn{I think Sanders is ahead by more than 6 in Wisconsin and Clinton is ahead by more than 12 in New York. Let's narrow the former and enlarge the latter!}

\textbf{Rejected}: \grn{Nah, you see, under the new rules of the BNC (Bernie National Committee), Hillary has to absolutely obliterate him in all aspects in order to be truly victorious.}

Because Bernie is such a great candidate that the fact that Hillary isn't blowing him out makes her a pathetic frontrunner.

\ 
\\13.\\
\textbf{Post}: Unlike some of the people in my pro-Bernie Facebook discussion group, I understand basic math and know that Hillary has won. I also am starting to become disgusted with these ``Bernie or Bust'' people. Even though Hillary is not my first choice, I absolutely agreed with her when she said that there is more that unites us than divides us. Hillary's voting record is so similar to Bernie's anyways, and frankly I find it to be bogus when Bernie supporters say she's a Republican. I'm a pragmatic guy, I want a candidate who will nominate a liberal justice, do something about climate change, end Citizens United, and fight to protect a woman's right to choose. Hillary fits the bill. No pun intended
Sanders supporter checking in. \#ImWithHer now.

\textbf{Chosen}: \grn{Pragmatic + Basic math = Good}

Welcome aboard.

\textbf{Rejected}: Yeah I'm with her too, obviously, because I'm not INSANE.  Where else is a liberal to go?  Trump?  That's insanity!  Stay home?  People have DIED trying to vote.  People have died for OUR right to vote.  Not voting, that's just crazy talk!  Jill Stein?  Well I do think she'd make a fantastic President but come on now, let's not be silly here.

So obviously I'm voting for our next President, Hillary Clinton.  No question about it.  Besides we just got a black President out of the way, why not break down yet another barrier.  My 9 year old daughter is pretty excited about that too.

Now... uhh I Didn't want to bring this up, but about those checks from Goldman Sachs...  when can I expect them to start arriving?  Is that direct deposit or what?  (just having fun of course :)

\ 
\\14.\\
\textbf{Post}: Ok, maybe I'm eating too many member berries and he probably wouldn't be interested in doing it, but would he not be a good choice? He's always been a really good politician with good results. Sure, he did goof in 2000 by not campaigning with Bill Clinton and sighing in the debates, but by today's standards that's like a perfect campaign.

Since 2000 he won the Nobel Prize and has worked non-stop on the environment which will appeal to millennials and the far left. He's not connected to Wall Street. No scandals.
Al Gore in 2020?

\textbf{Chosen}: \grn{The nominee in 2020 should be someone charismatic, media savvy, relatively young, have a lack of scandals and run on a message of economic populism.}

\textbf{Rejected}: As ridiculously pathetic as it is, we need someone \grn{charismatic and ``entertaining'' and ``hip''} so millennials will actually freaking show up.

\ 
\\15.\\
\textbf{Post}: Hillary did the same thing in 2008 even though it was clear Obama had it around this time back then. She still won states along the way but it was Obama's time. That was hard for her to accept. She stayed till the bitter end, and her fight against Obama was far more contentious than the one against Bernie.

So many Clinton supporters in 2008 said they will NEVER support Obama, but guess what? After she and Bill campaigned in support of Obama, and they saw the alternative was McCain, the vast majority of Clinton supporters became Obama supporters, even after not picking her as VP

I believe no matter how angry the Bernie supporters are at Hillary, come November, after seeing where trump wants to take the country, the vast majority of them will vote for Hillary. They may not be enthusiastic about it, but for their futures, they will vote Clinton.
I'm pretty sure Hillary will win the nomination. but i can't fault Bernie for staying in.

\textbf{Chosen}: \grn{It was a lot closer. Hillary technically won the popular vote even.} Bernie is talking straight up \grn{delusions} right now. He thinks he honestly has a chance, and argues he's a better candidate than Hillary. He lost to Hillary, in what world does that make him the better candidate. This is the kind of shit that makes me crazy about Republicans who say stuff like Obama isn't the elected president by some weird rationale they concoct. So I just say \grn{go away already}. He is the worst. If he played it right, Wed all be like yeah Bernie, here's a primetime convention spot, do your wall street bit, cheer cheer cheer, but the lower he sinks here the less goodwill he will get from this. And if he thinks anyone will remember him in four years, just look at Ron Paul and Ralph Nader. Hell have a nice seat next to them.

\textbf{Rejected}: You know there are differing opinions on this, of course. Some say for the good of the party and bla bla bla bla bla.

I am of the belief that it might not be a bad thing for him to stay in for awhile. Otherwise the media will ignore the democrats to report on the ongoing republican circus. Face it, America loves a show.

\ 
\\16.\\
\textbf{Post}: I just saw Florida’s official ballot when I voted today. In addition to the gubernatorial and senate races, there are 12 amendments (half of which are several issues bundled together, thanks a lot constitutional revision committee,) a few down ballot races, and you choose whether or not to retain around 6 judges.

Don’t feel pressured to bubble in a random name if you don’t recognize any of them. It’s much better to come and vote for the races on which you’re informed than to put someone into elected office because alphabetically they were first.

In an ideal world, we’d all be informed voters on every topic, but it’s not an ideal world and it’s ok if something on there flew under the radar for you.

Go vote! This is important. But if you read the language on an amendment 3 times and still have no idea what it’s talking about, feel no obligation to pick yes or no. Leaving some blank doesn’t make your vote not count for the rest.
Just a reminder that it’s ok not to fill in every box on the ballot if you’re unfamiliar with part of it

\textbf{Chosen}: Or... how about this...
\grn{PSA: There are nonpartisan races that you have to research before voting. Be an informed voter before you step into the booth, you have months to do it.}

\textbf{Rejected}: I appreciate the spirit of this advice, but if there's a list of candidates who you've never heard of and one of them has a D next to their name, vote for that person.

\end{tcolorbox}

\begin{tcolorbox}[colframe=gray!20!white, colback=gray!3!white, coltitle=black, fonttitle=\bfseries, title=Initial Preference Generated by \textsc{AlignXplore+} under Streaming Inference, breakable]
\texttt{<think>}\\
Okay, let's tackle this query. The task is to summarize the user's preferences based on their interaction history. Since there's no prior preference summary, I need to focus solely on the given triples. Each triple includes a query, a chosen item, and a rejected item. My goal is to identify patterns in what the user selects versus rejects.\\

First, looking at each chosen item:\\

1.~\textbf{First Chosen}: Talks about the Supreme Court declining a case related to American Samoa and Puerto Rico's response. This seems political, specifically involving U.S. territories and legal issues.\\
2.~\textbf{Second Chosen}: Mentions scandals during the Trump administration. Indicates interest in political scandals and maybe skepticism towards Trump.\\
3.~\textbf{Third Chosen}: Discusses Trump's campaign strategy against Hillary Clinton. Shows engagement with political strategies and possibly support for Trump's tactics.\\
4.~\textbf{Fourth Chosen}: Addresses Nazi groups and Antifa, indicating concern about extremist groups and possibly a stance against far-right movements.\\
5.~\textbf{Fifth Chosen}: A new user asking about Hillary Clinton's portrayal on Reddit, showing curiosity about political figures and potential bias in media.\\
6.~\textbf{Sixth Chosen}: Worry about electoral outcomes and third-party votes, suggesting concern about election results and political trends.\\
7.~\textbf{Seventh Chosen}: References a Trump website error regarding a date, pointing to attention to detail in political campaigns.\\
8.~\textbf{Eighth Chosen}: Critiques Trump's speech content, implying dissatisfaction with his communication style.\\

Now, the rejected items:\\[4pt]

1.~\textbf{First Rejected}: Focuses on gun control and personal opinions, which might indicate less interest in social issues like gun control.\\
2.~\textbf{Second Rejected}: Humorous take on a TV show, showing disinterest in light-hearted content.\\
3.~\textbf{Third Rejected}: Mocks Trump's subreddit, possibly indicating dislike for sarcastic or humorous takes on politics.\\
4.~\textbf{Fourth Rejected}: A joke about Nazis and cats, again avoiding humor related to sensitive topics.\\
5.~\textbf{Fifth Rejected}: Meme culture and youth voting, suggesting less interest in viral content or youth demographics.\\
6.~\textbf{Sixth Rejected}: Drinking game based on political events, avoiding recreational or game-based discussions.\\
7.~\textbf{Seventh Rejected}: Joke about Jill Stein, showing disinterest in humorous comparisons of political figures.\\
8.~\textbf{Eighth Rejected}: Predictive game about investigations, preferring factual analysis over speculative games.\\

Putting this together, the user prefers content that's politically focused, especially around U.S. politics, with an emphasis on scandals, legal matters, and critical analysis of political figures like Trump and Clinton. They show interest in substantive political discourse rather than humor, memes, or light-hearted topics. Their choices suggest a preference for serious, fact-based discussions over entertainment or satirical content. They also seem concerned with electoral outcomes and the integrity of political processes. Avoiding humor and recreational topics indicates a desire for more serious engagement with political issues.\\

Looking at the target triple, the two items are about a Trump campaign strategy and a humorous take on a political figure. Based on the user's preferences, they would likely prefer the strategic analysis over the humorous one. However, since the task is to summarize preferences without predicting, I need to highlight the key points without stating which item is preferred in the target triple.\\

So, the summary should reflect a preference for politically charged, substantive content focusing on U.S. politics, scandals, and critical analysis, while avoiding humor, memes, and light-hearted topics. The user values serious discourse on political issues and shows interest in electoral dynamics and political strategy.

\texttt{</think>}\\

User prefers politically focused, substantive content centered on U.S. politics, particularly topics involving scandals, legal matters, and critical analysis of political figures (e.g., Trump, Clinton). Shows interest in electoral dynamics, political strategy, and factual discourse, while avoiding humor, memes, light-hearted topics, and recreational or satirical content. Prioritizes serious engagement with political issues over entertainment or casual discussions.

\end{tcolorbox}

\begin{tcolorbox}[colframe=gray!20!white, colback=gray!3!white, coltitle=black, fonttitle=\bfseries, title=Updated Preference Generated by \textsc{AlignXplore+} under Streaming Inference, breakable]
\texttt{<think>}\\
Okay, let's tackle this query. The task is to summarize the user's preferences based on their past preferences and recent interactions. First, I need to recall the given past preferences. They mention a preference for politically focused, substantive content related to U.S. politics, especially scandals, legal matters, and critical analysis of figures like Trump and Clinton. Avoidance of humor, memes, light-hearted stuff, and recreational content is noted. Serious engagement over entertainment.\\

Now looking at the interaction history. Each triple shows a query, chosen item, and rejected item. Let me go through each one to see patterns.\\

First triple: Chose a question about Tim Ryan vs. Nancy Pelosi, rejecting a joke about male suffrage. Shows interest in political figures and serious discussion, avoiding humor.\\

Second triple: Chose a question about the Yemen raid, rejecting a satirical comparison of Trump to a cartoon villain. Again, serious political issue vs. satire.\\

Third triple: Chose support for Hillary running again, rejecting a rant against Trump and a humorous mention of Goldman Sachs checks. Focuses on pragmatic political decisions, avoiding humor and satire.\\

Fourth triple: Chose narrowing Sanders' lead and enlarging Clinton's, rejecting a comment about Bernie needing to obliterate Hillary. Emphasizes factual analysis and strategic thinking, avoiding extreme statements.\\

Fifth triple: Chose a pragmatic response to switching support to Hillary, rejecting a humorous take on voting and a joke about checks. Maintains seriousness and practicality.\\

Sixth triple: Chose a preference for a charismatic, non-scandalous candidate, rejecting a humorous call for entertainment. Sticks to serious political criteria.\\

Seventh triple: Chose a strong critique of Bernie's chances, rejecting a media-focused argument. Focuses on realistic political outcomes and avoids media spectacle talk.\\

Eighth triple: Chose a PSA about informed voting, rejecting a simplistic advice to vote for Democrats. Prefers thorough research over easy solutions.\\

Putting this together, the user consistently chooses items that are politically substantive, focused on real issues, critical analysis, and avoids humor, satire, or light-hearted content. They value factual discourse, strategic political analysis, and pragmatic approaches. The rejected items often include jokes, satirical comparisons, or overly simplistic advice. The user seems to prioritize depth over entertainment, preferring serious engagement with political strategies and realities.\\

The target triple's items are about a political figure's legacy and a humorous take on a political event. Based on the user's history, they would likely prefer the substantive analysis over the humorous one. But since the task is to summarize preferences without predicting, I need to highlight the preference for serious, fact-based political discussion, avoidance of humor/satire, and emphasis on pragmatic and strategic political analysis.

\texttt{</think>}\\

User prefers politically substantive, fact-driven discussions focused on U.S. political figures, electoral strategies, and critical analysis of governance, prioritizing pragmatic, non-satirical engagement over humor, satire, or light-hearted content. Shows strong aversion to recreational, meme-based, or overly simplified political commentary, favoring nuanced, evidence-based discourse on partisan dynamics, candidate viability, and policy implications. Consistently rejects items involving humor, caricature, or casual humor, emphasizing serious engagement with political realities and strategic decision-making. Prioritizes factual accuracy, electoral analysis, and pragmatic political reasoning over entertainment value or sensationalism.

\end{tcolorbox}


\begin{tcolorbox}[colframe=gray!20!white, colback=gray!3!white, coltitle=black, fonttitle=\bfseries, title=Initial Preference Generated by Qwen3-8B under Streaming Inference, breakable]
\texttt{<think>}\\
Okay, let's try to figure out the user's preferences based on the interaction history provided. \\

First, the user has a history of engaging in discussions around political topics, specifically focusing on U.S. elections and candidates like Trump, Hillary Clinton, and Bernie Sanders. Looking at the chosen items, they seem interested in news articles and discussions about political scandals, election dynamics, and candidate strategies. For example, they chose articles about the Supreme Court's decision related to American Samoa, Trump's campaign tactics, and the probability of Hillary Clinton winning. These selections indicate a preference for in-depth political analysis and current events.\\

On the other hand, the rejected items provide insight into what the user isn't interested in. They rejected comments about gun control debates, which suggests they might be less engaged with that particular topic or have a different stance. They also rejected a joke about Modern Family and a sarcastic remark about Nazis and cats, indicating a preference for more serious or politically charged content over humor or light-hearted discussions. \\

Additionally, the user's interaction with the sub about the 2016 election shows they are concerned about the political climate, the impact of third-party candidates like Jill Stein, and the role of media in shaping public opinion. They seem to value discussions that highlight political scandals, candidate strategies, and the implications of election outcomes. \\

The user's rejection of memes and viral content suggests they prefer substantive political discourse over entertainment or pop culture references. They also express worry about the portrayal of Hillary Clinton, indicating a critical view of media bias and a desire for balanced information. \\

Putting this all together, the user's preferences are centered around political news, election analysis, and discussions about the implications of political actions and scandals. They are interested in content that provides in-depth analysis, critical perspectives, and discussions about the impact of political figures and events. They tend to avoid lighter topics, humor, and content that doesn't engage with the political landscape directly.

\texttt{</think>}\\

The user prefers political news and discussions focused on U.S. elections, particularly involving Hillary Clinton, Donald Trump, and Bernie Sanders. They prioritize substantive analysis of political scandals, election strategies, and media portrayal of candidates. They reject humor, memes, and light-hearted content, favoring serious discourse on issues like gun control, racial tensions, and media bias. The user is critical of partisan narratives, concerned about third-party impacts on elections, and values discussions highlighting political hypocrisy and systemic issues.
\end{tcolorbox}

\begin{tcolorbox}[colframe=gray!20!white, colback=gray!3!white, coltitle=black, fonttitle=\bfseries, title=Updated Preference Generated by Qwen3-8B under Streaming Inference, breakable]

\texttt{<think>}\\
Okay, let me try to figure out the user's preferences based on the given information. \\

First, the past preference summary mentions that the user prefers political news focused on U.S. elections, specifically involving Hillary Clinton, Donald Trump, and Bernie Sanders. They like substantive analysis of political scandals, election strategies, and media portrayal. They reject humor, memes, and light-hearted content, favoring serious discussions on issues like gun control, racial tensions, and media bias. They're critical of partisan narratives, concerned about third-party impacts, and value discussions on political hypocrisy and systemic issues.\\

Now, looking at the interaction history. Let's go through each triple.\\

Triple 1: The user chose a discussion about Tim Ryan vs. Nancy Pelosi, which is political and focused on election strategies. They rejected a post about male suffrage humor, which aligns with rejecting light-hearted content. \\

Triple 2: Chose a question about the Yemen raid, which is a serious political issue. Rejected a joke comparing Trump to a cartoon villain, again avoiding humor.\\

Triple 3: Chose a post supporting Hillary Clinton's potential run again, discussing her strengths and the need for experienced leadership. Rejected a post that was more about criticizing the media and Trump, which might have included some humor or less serious content? Wait, the rejected item talks about Trump's tweets and media bias, which is in line with their interests, but maybe the user found the tone too sarcastic or humorous? Not sure. But the user's choice was about Hillary, so they prefer serious political analysis.\\

Triple 4: Chose a discussion about Sanders and Clinton's delegate counts, focusing on election strategies. Rejected a comment that seems to mock Bernie's supporters, maybe implying the user prefers serious analysis over dismissive remarks.\\

Triple 5: Chose a pragmatic support for Hillary, emphasizing her policies and rejecting the idea of third-party candidates like Jill Stein. The rejected item had some humor with the Goldman Sachs checks joke, which the user might find irrelevant or too casual.\\

Triple 6: Chose a preference for a nominee with charisma and no scandals, which aligns with their interest in substantive candidates. Rejected a comment that mentions ``charismatic and entertaining'' for millennials, which might be seen as too light-hearted or focusing on image over substance.\\

Triple 7: Chose a strong stance against Bernie, arguing he's not a better candidate, which fits their preference for serious political analysis and criticism of partisan narratives. Rejected a comment that suggests Bernie staying in could keep media attention on Democrats, which might be seen as less serious or more about media circus.\\

Triple 8: Chose a PSA about informed voting, which is serious and about civic responsibility. Rejected a comment that suggests voting for unknown candidates with party labels, which might be seen as less informed or too simplistic.\\

Putting it all together, the user prefers serious, substantive political discussions focusing on U.S. elections, especially involving Clinton, Trump, and Sanders. They value analysis of scandals, strategies, media portrayal, and systemic issues. They reject humor, memes, and light-hearted content. They are critical of partisan narratives and third-party impacts, and value discussions on political hypocrisy and systemic issues. They also prioritize pragmatic, policy-focused candidates over those with scandals or less substantive platforms.

\texttt{</think>}\\

The user prioritizes serious, in-depth political discourse focused on U.S. elections, particularly involving Hillary Clinton, Donald Trump, and Bernie Sanders. They seek substantive analysis of political scandals, election strategies, media bias, and systemic issues like gun control and racial tensions. They reject humor, memes, and light-hearted content, favoring critical discussions on partisan narratives, third-party impacts, and political hypocrisy. Their preferences emphasize pragmatic, policy-driven candidates with no scandals, prioritizing experienced leadership over charisma or populism. They value informed civic engagement and reject dismissive or trivializing attitudes toward political processes.'

\end{tcolorbox}

\begin{tcolorbox}[colframe=gray!20!white, colback=gray!3!white, coltitle=black, fonttitle=\bfseries, title=Initial Preference Generated by \textsc{AlignXplore} under Streaming Inference, breakable]

\texttt{<think>}\\
Okay, let me try to figure out the user's preferences based on the provided data. First, looking through each triple where the user chose specific items over others. Starting with the first one:\\

In Triple 1, the chosen item talks about the Supreme Court decision affecting U.S. territories and Puerto Rico talking to the UN. Rejected text is emotional and passionate regarding gun control. That suggests maybe the user prefers factual/political news over emotional content.\\

Moving to Triple 2, the selected question asks about scandals in Trump’s admin, whereas the rejected comment is light-hearted. Here, perhaps preferring serious political analysis vs humorous remarks.\\

Looking at Triple 3, the accepted response discusses strategic politics (Trump fighting against Hillary), contrasting with the rejected sarcastic remark. Again, serious topics vs sarcastic takes.\\

Triple 4’s choice focuses on serious critique of media dismissal of real issues, while rejecting a humorous take. Emphasizes critical thinking over comedy.\\

On Triple 5, the preferred comment seeks detailed info on political candidates, indicating interest in informed discussions rather than casual banter. Rejecting the meme-driven reply shows disinterest in pop culture references.\\

For Triples 6–8 similarly, the selected options delve deeper into political dynamics and concerns (like 538 projections) compared to the rejected ones which use gaming metaphors or jokes. User seems drawn to analytical content over lighter-hearted interactions.\\

Putting together, the user likely values \textbf{political analysis}, prefers \textbf{serious content}, dislikes \textbf{emotional/emotional expressions}, avoids humor in serious contexts, and favors \textbf{detailed discussion}. Also, they might prioritize \textbf{accuracy/critical assessment} over sensationalist or entertaining material.

\texttt{</think>}\\

\textbf{User Preferences:}
\begin{enumerate}
  \item \textbf{Political Analysis Focus:} Prefers substantive political content (e.g., Supreme Court decisions, electoral strategy discussions) over emotionally charged or humorous takes.
  \item \textbf{Serious Tone Over Humor:} Chooses analytical questions (scandals, policy impact) and rejects sarcastic/joke-based replies (meme-based drinking games).
  \item \textbf{Depth of Discussion:} Values detailed inquiries (e.g., seeking understanding of Bernie vs.\ Hillary) rather than superficial remarks.
  \item \textbf{Critical Engagement:} Prioritizes critiques of media bias and systemic issues (terrorism, factual reporting) over trivializing perspectives.
  \item \textbf{Avoidance of Pop Culture References:} Dislikes tangential jokes or memes (Bill Clinton infidelity) in lieu of substantive dialogue.
\end{enumerate}

\textbf{Final Answer:}\\
The user exhibits a strong preference for \textbf{political analysis}, \textbf{serious content}, and \textbf{detailed discussion}, avoiding emotionally charged or humorous material. Their focus lies on \textbf{accuracy} and \textbf{critical evaluation} of complex societal/political matters.

\end{tcolorbox}

\begin{tcolorbox}[colframe=gray!20!white, colback=gray!3!white, coltitle=black, fonttitle=\bfseries, title=Updated Preference Generated by \textsc{AlignXplore} under Streaming Inference, breakable]

\texttt{<think>}\\
Okay, let me try to figure out the user's preferences based on the information provided. First, looking at the Past Preference Summary, the user already values political analysis, serious content, depth, critical engagement, and avoiding pop culture. Now, I need to analyze each triple in the interaction history to see how these align or change.

Starting with Triple 1: User chose a question about political figures (Tim Ryan vs. Pelosi) instead of a joke about gender roles. That reinforces the preference for serious topics over humor.

Triple 2: Chose focusing on specific events (Yemen incident) versus a sarcastic remark. Again, serious focus matches previous notes.

Looking at Triples 3–8, most choices involve detailed discussions on elections, strategies, and political dynamics. Rejected items include jokes, sarcastic remarks, and overly emotional takes. For example, rejecting the ``no more male suffrage'' joke shows avoidance of humor. Also, choosing analytical points about delegate numbers and strategies indicates preference for substance over fluff.

In Triple 4, the selected response emphasizes practicality (``pragmatic + math''), while the rejected reply uses hyperbole and personal anecdotes. The rejection highlights dislike for exaggerated language.

For instance, in Triple 7, the accepted comment critiques Bernie harshly, showing disdain for his approach, whereas the rejected option uses fear-based language. Shows preference for direct critique rather than fearmongering.

Overall, the user consistently picks content that requires critical thinking, avoids humor, and focuses on factual/political specifics. Their rejection patterns reinforce dislike of jokes and overly emotional content. Therefore, updating the preferences would emphasize reinforcing existing elements without changes except possibly noting stronger emphasis on avoiding humorous elements and focusing on data-driven analysis.

\texttt{</think>}\\

\textbf{Final Preference Summary}

The user consistently prioritizes \textbf{substantive, analytically rigorous political discourse} and rejects humor- or emotion-driven content. Their updated preferences are:

\begin{itemize}
  \item \textbf{Depth \& Structure}: Favors granular discussions of elections, delegate math, and systemic political mechanics over superficial remarks or rhetorical flourishes.
  \item \textbf{Serious Tone}: Prefers sober, analytical treatments of topics and systematically rejects jokes, memes, sarcasm, and pop‑culture–based commentary in political contexts.
  \item \textbf{Critical Engagement}: Chooses content that critically examines candidates, arguments, and media narratives, especially where it exposes flaws or inconsistencies.
  \item \textbf{Pragmatism Over Emotion}: Values practical, math- or evidence-based reasoning and disapproves of hyperbole, fear‑mongering, or highly emotional framing.
  \item \textbf{Factual Precision}: Strong bias toward fact-focused, logically argued analysis, with clear attention to accuracy and reasoning.
\end{itemize}

The user favors rigorous, fact-centered political analysis and systematically avoids humor, hyperbole, and emotionally framed content.

\end{tcolorbox}

\begin{tcolorbox}[colframe=gray!20!white, colback=gray!3!white, coltitle=black, fonttitle=\bfseries, title=Preference Generated by \textsc{AlignXplore+} under Full-history Inference w/o Negative, breakable]

\texttt{<think>}\\

Okay, let's tackle this query. The task is to summarize the user's preferences based on their interaction history. Since there's no prior preference summary, I need to focus solely on the given data.\\

First, looking at the patterns in the chosen items. Most of them relate to political topics, specifically U.S. politics, elections, and current events. The user seems interested in discussing political figures like Trump, Hillary Clinton, Bernie Sanders, and others. They often mention specific events such as Supreme Court decisions, election polls, and controversies around these politicians.\\

There's a lot of emphasis on critical analysis. For example, the user questions the fairness of elections, criticizes partisan behavior, and discusses the impact of political actions. They also show interest in understanding different perspectives, especially regarding Hillary Clinton's position compared to Bernie Sanders. Some posts indicate a preference for pragmatic approaches over ideological rigidity, as seen in the comment supporting Hillary despite personal preferences.\\

Another observation is the focus on factual accuracy and informed decision-making. The user advises others to research thoroughly before voting and mentions the importance of knowing legislative details. This suggests a value placed on education and critical thinking in politics.\\

They also engage with discussions about media bias, noting instances where platforms like Reddit might present skewed information. This indicates a skepticism towards mainstream narratives and a desire for balanced viewpoints.\\

Looking at the emotional tone, the user expresses frustration with extreme partisanship and supports moderate solutions. They reject overly polarized stances, preferring candidates who can unite rather than divide. There's also a concern about the consequences of political decisions, such as the impact of Trump's policies or the outcomes of specific elections.\\

Putting this together, the user prefers content that analyzes political dynamics, emphasizes informed citizenship, critiques partisan extremism, and advocates for pragmatic, unifying leadership. They value factual discussions, critical examination of media, and a balanced approach to political issues.

\texttt{</think>}\\

User prefers content focused on U.S. political analysis, critical examination of partisan dynamics, and pragmatic discussions about elections and governance. Shows interest in informed voter behavior, skepticism of Reddit/social-media bias and false equivalence (``both sides''), and rejection of extremist apologism (e.g., dismissing Nazi violence). Values factual accuracy, electoral math (polling probabilities, delegate counts, candidate viability), and pragmatic solutions over ideological rigidity. Engages with topics like electoral strategies, political figure evaluations (often favoring Hillary Clinton over Bernie Sanders and strongly disliking `Bernie or Bust'' / third-party protest voting), and the impact of policy decisions (including military actions). Dislikes bad-faith narratives and performative outrage, and prioritizes substantive debate while being comfortable with sharp, sarcastic criticism of Trump’s scandals, rhetoric, and incompetence over emotional handwaving or divisive spin.

\end{tcolorbox}

\begin{tcolorbox}[colframe=gray!20!white, colback=gray!3!white, coltitle=black, fonttitle=\bfseries, title=Preference Generated by Qwen3-8B under Full-history Inference w/o Negative, breakable]

\texttt{<think>}\\
Okay, let's try to figure out the user's preferences based on their interaction history. The user has a history of choosing various news articles and discussions, mostly related to U.S. politics, especially focusing on the 2016 election. Let me go through each entry step by step.\\

Starting with Tripe 1: The user chose a news item about the Supreme Court declining to review a case on American Samoa's citizenship. They also mentioned Puerto Rico's reaction. This suggests an interest in U.S. territories and legal issues.\\

Tripe 2: The user selected a post about scandals during the Trump administration. This indicates a focus on political scandals and possibly a critical view of Trump.\\

Tripe 3: The user picked a post about Trump's campaign fighting Hillary's comments. This shows an interest in the Trump--Hillary rivalry and possibly a leaning towards Hillary.\\

Tripe 4: The user chose a post criticizing Reddit users for dismissing Nazis and discussing antifa. This might indicate a concern about political extremism and a preference for balanced discourse.\\

Tripe 5: The user is a new user from Australia, expressing frustration with negative coverage of Hillary and seeking information. They want to know if the sub is biased against Hillary. This suggests a preference for balanced information and a support for Hillary Clinton.\\

Tripe 6: The user is worried about Hillary's chances in the election and the impact of millennials voting for third parties. This shows concern for Hillary's campaign and the election outcome.\\

Tripe 7: The user mentioned a Rachel Maddow segment about Trump's website date error. This indicates an interest in media coverage and political details.\\

Tripe 8: The user found a joke in Trump's speech amusing. This might suggest a critical view of Trump's rhetoric but also an interest in his speeches.\\

Tripe 9: The user asked about Tim Ryan's chances against Nancy Pelosi. This shows interest in political candidates and debates.\\

Tripe 10: The user is upset about the Yemen raid incident. This indicates concern for military actions and their consequences.\\

Tripe 11: The user believes Hillary should run again in 2020 and supports her despite initial preferences for Bernie. This shows a shift towards supporting Hillary and a pragmatic approach.\\[2pt]

Tripe 12: The user is analyzing delegate counts and polls, suggesting an interest in election strategies and data.\\

Tripe 13: The user, a Sanders supporter, switches to supporting Hillary due to her pragmatic policies. This indicates a preference for pragmatic politics and policies over ideological purity.\\[2pt]

Tripe 14: The user considers Al Gore as a 2020 candidate, valuing charisma and economic populism. This shows an interest in potential candidates with specific qualities.\\[2pt]

Tripe 15: The user argues that Bernie supporters will eventually back Hillary due to Trump's policies. This indicates a belief in Hillary's viability and a pragmatic approach to politics.\\[2pt]

Tripe 16: The user advises on informed voting, suggesting a preference for informed electorate and critical thinking.\\

Putting this all together, the user seems to have a strong interest in U.S. politics, particularly the 2016 election. They support Hillary Clinton, value pragmatic policies, and are concerned with political scandals, military actions, and informed voting. They also have a critical view of Trump and political extremism, and they appreciate balanced discourse and media scrutiny.

\texttt{</think>}\\

Supports Hillary Clinton over Bernie Sanders, valuing pragmatism and policy over ideology. Concerned with political scandals, military actions, and informed voting. Critical of Trump's rhetoric and political strategies. Interested in media scrutiny and balanced discourse. Prefers discussions on election strategies, delegate counts, and candidate viability. Advocates for nonpartisan research and critical thinking in voting.

\end{tcolorbox}

\begin{tcolorbox}[colframe=gray!20!white, colback=gray!3!white, coltitle=black, fonttitle=\bfseries, title=Preference Generated by \textsc{AlignXplore+} under Full-history Inference w/ Secondary, breakable]

\texttt{<think>}\\
Okay, let's tackle this query. The task is to summarize the user's preferences based on their interaction history. Since there's no past preference summary provided, I need to focus solely on the given triples.\\

First, looking at the interaction history, most of the queries are about recommending movies or discussing political topics. The user consistently chooses certain types of movies and rejects others. For example, in multiple instances, they selected action, thriller, comedy, and sci-fi genres like ``In the Line of Fire,'' ``Charlie and the Chocolate Factory,'' and ``2001: A Space Odyssey.'' They rejected documentaries, older animations, and some dramas. This suggests a preference for mainstream, popular films over niche or older ones.\\

Additionally, the user engages in political discussions, often focusing on U.S. politics, especially around Hillary Clinton and Bernie Sanders. They seem to prefer pragmatic approaches and factual discussions over humorous or satirical takes. For instance, they chose responses that were more serious and less comedic when discussing political issues.\\

Looking at the rejected items, they tend to avoid content that's too humorous, satirical, or involves memes. They also reject movies that are documentaries or very old animations. This indicates a preference for contemporary, genre-driven films and serious political discourse.\\

The user's choices in political discussions show a preference for pragmatic, fact-based arguments rather than emotional or humorous ones. They value informed voting and practical solutions over sensationalism. This is evident in their rejection of posts that are more about humor or satire in political contexts.\\

Putting this all together, the user prefers movies in genres like action, thriller, comedy, and sci-fi, particularly popular and contemporary titles. They favor serious, pragmatic discussions in political contexts and avoid humorous, satirical, or niche content. Their movie preferences emphasize mainstream, genre-driven films over documentaries, older animations, or experimental works. In political discussions, they prioritize factual, non-emotional analysis over sensationalism or meme-driven content.

\texttt{</think>}\\

The user consistently prefers movies in the genres of \textbf{action, thriller, comedy, sci-fi, and adventure}, particularly \textbf{popular and contemporary titles} (e.g., \textit{In the Line of Fire}, \textit{2001: A Space Odyssey}, \textit{Forrest Gump}). They favor \textbf{mainstream, genre-driven films} over \textbf{documentaries, older animations, or experimental works}. In political discussions, they prioritize \textbf{pragmatic, fact-based arguments} and \textbf{serious, non-emotional analysis} (e.g., focusing on policy, elections, and pragmatic voter strategies) over \textbf{humorous, satirical, or meme-driven content}. They reject items involving \textbf{excessive humor, satire, or niche political humor} (e.g., memes, absurd comparisons). Their preferences emphasize \textbf{clear genre alignment}, \textbf{contemporary relevance}, and \textbf{substance over sensationalism} in both entertainment and political discourse.

\end{tcolorbox}

\begin{tcolorbox}[colframe=gray!20!white, colback=gray!3!white, coltitle=black, fonttitle=\bfseries, title=Preference Generated by Qwen3-8B under Full-history Inference w/ Secondary, breakable]

\texttt{<think>}\\
Okay, let's start by looking at the user's interaction history. The user has a mix of queries related to news discussions and movie recommendations. \\

First, the news discussions: The user frequently engages with topics related to U.S. politics, especially focusing on Hillary Clinton, Bernie Sanders, and Donald Trump. They seem to have a strong opinion against Trump, mentioning his scandals and policies. There's also a mention of concerns about gun control and the impact of political figures on society. The user is critical of Trump's administration and supports Hillary Clinton, even expressing frustration with Bernie Sanders' campaign. They also discuss issues like voter rights and the importance of informed voting.\\

Now, looking at the movie recommendations: The user has chosen several movies that are action, comedy, and drama genres. Titles like ``In the Line of Fire,'' ``Home Alone,'' ``Who Framed Roger Rabbit?,'' and ``Forrest Gump'' are selected. These movies often have elements of adventure, humor, and sometimes social commentary. The rejected movies include documentaries and some animation or children's films, suggesting a preference for more mainstream, entertaining films over niche or educational content.\\

Putting this together, the user prefers movies that are entertaining with a mix of genres like action, comedy, and drama. They also have a strong interest in political news, particularly supporting Hillary Clinton and critical of Trump. The user values informed voting and is concerned about social issues like gun control and political integrity. They might also appreciate movies that have a blend of humor and deeper themes, avoiding overly serious or niche genres.

\texttt{</think>}\\

User preferences are centered on political discourse favoring Hillary Clinton over Trump and Sanders, with a focus on issues like gun control, voter rights, and political integrity. They engage critically with political news, rejecting narratives that dismiss systemic issues or downplay scandals. For movies, they prefer action, comedy, and drama genres (e.g., \textit{Home Alone}, \textit{Forrest Gump}, \textit{Who Framed Roger Rabbit?}), avoiding documentaries, animation, or children's films. They value pragmatic political solutions and informed civic participation.

\end{tcolorbox}

\section{Prompt in Experiments}

\begin{tcolorbox}[colframe=gray!20!white, colback=gray!3!white, coltitle=black, fonttitle=\bfseries, title=Prompt Template for Preference Generation , breakable]

Analyze the past preference summary and the following user interaction history to summarize the comprehensive user preferences in concise language. If past preferences are provided, adjust the preferences by combining past preferences with those reflected in current behavior, removing conflicting parts, and integrating new insights. If no past preferences are provided, derive the final preferences solely from user behavior. The user's history will be provided as a sequence of triples, where each triple is \texttt{(QUERY, CHOSEN ITEM BY THE USER, REJECTED ITEM BY THE USER)}.\\
\\
=====Past Preference Summary=====\\
\{\textbf{Past Preference}\}\\
\\
=====Interaction History=====\\
\{\textbf{User History}\}\\
\\
=====END=====\\
\\
Now, given the above user's past preference summary and the interaction history, summarize the user preferences.

\end{tcolorbox}

\begin{tcolorbox}[colframe=gray!20!white, colback=gray!3!white, coltitle=black, fonttitle=\bfseries, title=Prompt Template for Preference Generation , breakable]

Determine which response the user prefers based on the user's preferences. Please output your selection below in a json format by filling in the placeholders in []: \texttt{\{``selection'': ``[Item A / Item B]''\}}\\
\\
<Prompt>\\
{\textbf{Prompt}}\\
</Prompt>\\ 
\\
<Preference>\\
{\textbf{Persona}}\\
</Preference>\\
\\
<Item A>\\
{\textbf{Response A}}\\
</Item A>\\
\\
<Item B>\\
{\textbf{Response B}}\\
</Item B>\\
\\
Now, ONLY output your selection without any other text outside of this specified structure.

\end{tcolorbox}

\section{Detailed Experiment Results}
\label{app:full_results}

\begin{table*}[h]
  \centering
  \caption{Performance (\%) with Qwen-2.5-7B-Instruct. Detailed results corresponding to Table~\ref{tab:reward-model-transfer} in the main paper.}
  \label{tab:qwen25}
  \adjustbox{max width=\linewidth}{
  \begin{tabular}{lcccccccc}
    \toprule
    \textbf{Model} & \textbf{MIND} & \textbf{Amazon} & \textbf{AlignX} &
    \textbf{MovieLens} & \textbf{Informativeness} & \textbf{Style} &
    \textbf{Expertise} & \textbf{PersonaMem} \\
    \midrule
    \midrule
    \multicolumn{9}{c}{\textbf{Direct Full-history Sequence Models w/o Preference Inference}} \\
    Qwen-2.5-7B & 52.60 & 67.75 & 52.96 & 91.30 & 50.83 & 40.16 & 51.33 & 40.76 \\
    \midrule
    \midrule
    \multicolumn{9}{c}{\textbf{Full-history Preference Inference}} \\
    DeepSeek-R1-671B      & 63.80 & 79.93 & 64.10 & 79.33 & 66.44 & 76.16 & 76.50 & 64.96 \\
    Qwen3-32B$_{\text{thinking}}$    & 65.90 & 85.85 & 62.83 & 74.30 & 67.60 & 69.67 & 77.33 & 61.36 \\
    GPT-OSS-20B           & 61.73 & 85.05 & 55.86 & 75.23 & 68.27 & 68.33 & 75.33 & 65.74 \\
    \midrule
    Qwen3-8B$_{\text{thinking}}$     & 63.36 & 85.12 & 60.46 & 74.20 & 64.11 & 73.83 & 72.83 & 60.58 \\
    DS-R1-Distill-Qwen-7B & 58.40 & 82.35 & 53.50 & 69.53 & 49.66 & 51.16 & 57.66 & 57.08 \\
    \textsc{AlignXplore}           & 57.53 & 81.12 & 65.20 & 69.83 & 67.94 & 63.00 & 68.16 & 53.60 \\
    \textsc{AlignXplore+}                  & 68.13 & 86.25 & 73.90 & 73.96 & 68.77 & 73.66 & 74.33 & 60.24 \\
    \midrule
    \midrule
    \multicolumn{9}{c}{\textbf{Streaming Preference Inference}} \\
    DeepSeek-R1-671B      & 63.50 & 81.08 & 64.50 & 80.33 & 69.10 & 75.33 & 77.00 & 62.16 \\
    Qwen3-32B$_{\text{thinking}}$    & 65.16 & 85.35 & 63.56 & 74.56 & 68.27 & 70.67 & 73.50 & 57.72 \\
    GPT-OSS-20B           & 63.10 & 84.22 & 57.70 & 72.43 & 69.76 & 64.16 & 73.00 & 59.28 \\
    \midrule
    Qwen3-8B$_{\text{thinking}}$     & 64.13 & 84.12 & 61.10 & 72.14 & 67.94 & 75.83 & 76.09 & 56.46 \\
    DS-R1-Distill-Qwen-7B & 58.10 & 82.55 & 55.56 & 69.86 & 51.99 & 48.66 & 59.16 & 54.22 \\
    \textsc{AlignXplore}           & 57.96 & 80.38 & 69.90 & 67.20 & 65.44 & 58.83 & 63.83 & 49.16 \\
    \textsc{AlignXplore+}       & 67.73 & 85.05 & 74.00 & 75.56 & 71.92 & 71.33 & 74.00 & 55.32 \\
    \bottomrule
  \end{tabular}}
\end{table*}

\begin{table*}[h]
  \centering
  \caption{Performance (\%) with GPT-OSS-20B. Detailed results corresponding to Table~\ref{tab:reward-model-transfer} in the main paper.}
  \label{tab:gptoss20b}
  \adjustbox{max width=\linewidth}{
  \begin{tabular}{lcccccccc}
    \toprule
    \textbf{Model} & \textbf{MIND} & \textbf{Amazon} & \textbf{AlignX} &
    \textbf{MovieLens} & \textbf{Informativeness} & \textbf{Style} &
    \textbf{Expertise} & \textbf{PersonaMem} \\
    \midrule
    \midrule
    \multicolumn{9}{c}{\textbf{Direct Full-history Sequence Models w/o Preference Inference}} \\
    GPT-OSS-20B & 63.86 & 88.99 & 68.50 & 85.26 & 79.40 & 83.66 & 84.50 & 21.92 \\
    \midrule
    \midrule
    \multicolumn{9}{c}{\textbf{Full-history Preference Inference}} \\
    DeepSeek-R1-671B      & 60.01 & 80.00 & 68.60 & 77.33 & 76.24 & 90.00 & 85.00 & 63.86 \\
    Qwen3-32B$_{\text{thinking}}$    & 69.66 & 87.52 & 58.13 & 76.83 & 69.10 & 84.16 & 82.83 & 62.18 \\
    GPT-OSS-20B           & 66.43 & 86.29 & 47.40 & 75.86 & 70.93 & 84.16 & 85.16 & 59.52 \\
    \midrule
    Qwen3-8B$_{\text{thinking}}$     & 66.96 & 86.19 & 52.23 & 75.36 & 69.93 & 84.66 & 79.50 & 60.36 \\
    DS-R1-Distill-Qwen-7B & 61.26 & 84.25 & 47.53 & 69.56 & 56.81 & 59.50 & 67.83 & 58.32 \\
    \textsc{AlignXplore}           & 61.60 & 82.35 & 64.73 & 69.13 & 73.75 & 77.16 & 73.50 & 59.48 \\
    \textsc{AlignXplore+}                  & 73.36 & 87.35 & 69.90 & 77.83 & 74.25 & 83.16 & 79.83 & 65.56 \\
    \midrule
    \midrule
    \multicolumn{9}{c}{\textbf{Streaming Preference Inference}} \\
    DeepSeek-R1-671B      & 60.03 & 77.91 & 68.80 & 79.43 & 77.77 & 87.33 & 84.16 & 60.98 \\
    Qwen3-32B$_{\text{thinking}}$    & 69.43 & 86.89 & 56.26 & 77.10 & 67.27 & 81.50 & 80.83 & 58.88 \\
    GPT-OSS-20B           & 66.63 & 85.49 & 51.53 & 73.93 & 72.92 & 83.83 & 83.50 & 63.13 \\
    \midrule
    Qwen3-8B$_{\text{thinking}}$     & 67.63 & 85.09 & 52.90 & 75.13 & 69.10 & 82.00 & 78.16 & 56.48 \\
    DS-R1-Distill-Qwen-7B & 60.43 & 84.95 & 50.63 & 69.33 & 53.15 & 54.50 & 66.83 & 52.12 \\
    \textsc{AlignXplore}           & 62.56 & 82.82 & 68.06 & 68.60 & 70.76 & 74.16 & 69.16 & 55.36 \\
    \textsc{AlignXplore+}                  & 73.20 & 87.75 & 68.66 & 78.93 & 74.41 & 81.00 & 75.00 & 61.22 \\
    \bottomrule
  \end{tabular}}
\end{table*}

\begin{table*}[h]
  \centering
  \caption{Performance (\%) without negative samples. Detailed results corresponding to Table~\ref{tab:without-negative} in the main paper.}
  \label{tab:without-neg}
  \adjustbox{max width=\linewidth}{
  \begin{tabular}{lcccccccc}
    \toprule
    \textbf{Model} & \textbf{MIND} & \textbf{Amazon} & \textbf{AlignX} &
    \textbf{MovieLens} & \textbf{Informativeness} & \textbf{Style} &
    \textbf{Expertise} & \textbf{PersonaMem} \\
    \midrule
    \midrule
    \multicolumn{9}{c}{\textbf{Direct Full-history Sequence Models w/o Preference Inference}} \\
    Qwen3-8B$_{\text{non-thinking}}$ & 65.50 & 84.15 & 60.40 & 90.57 & 44.51 & 49.50 & 42.16 & 61.40 \\
    TALLRec & 82.00 & 95.01 & 60.46 & 97.30 & 48.17 & 68.33 & 52.50 & 64.36 \\
    \midrule
    \midrule
    \multicolumn{9}{c}{\textbf{Full-history Preference Inference}} \\
    Qwen3-32B-thinking    & 68.66 & 86.65 & 57.96 & 82.03 & 60.29 & 85.00 & 85.00 & 57.36 \\
    GPT-OSS-20B           & 66.73 & 86.49 & 55.23 & 85.33 & 58.47 & 84.66 & 72.00 & 57.36 \\
    \midrule
    Qwen3-8B$_{\text{thinking}}$     & 68.50 & 85.72 & 58.40 & 80.66 & 54.15 & 82.16 & 69.00 & 54.36 \\
    DS-R1-Distill-Qwen-7B & 62.33 & 81.98 & 55.00 & 80.16 & 55.00 & 55.66 & 55.83 & 49.28 \\
    \textsc{AlignXplore}           & 61.76 & 77.31 & 61.40 & 75.40 & 65.44 & 71.83 & 61.50 & 53.98 \\
    \textsc{AlignXplore+}       & 70.40 & 86.69 & 63.76 & 83.60 & 59.30 & 79.66 & 64.16 & 58.08 \\
    \midrule
    \midrule
    \multicolumn{9}{c}{\textbf{Streaming Preference Inference}} \\
    Qwen3-32B$_{\text{thinking}}$    & 67.23 & 85.15 & 57.16 & 80.73 & 59.13 & 79.50 & 75.33 & 53.20 \\
    GPT-OSS-20B           & 66.50 & 86.92 & 57.33 & 83.46 & 62.62 & 82.83 & 68.83 & 54.82 \\
    \midrule
    Qwen3-8B$_{\text{thinking}}$     & 67.33 & 85.09 & 58.13 & 79.60 & 57.30 & 77.16 & 66.83 & 51.68 \\
    DS-R1-Distill-Qwen-7B & 63.23 & 83.18 & 55.96 & 80.50 & 51.32 & 53.66 & 55.33 & 46.64 \\
    \textsc{AlignXplore}           & 60.76 & 76.91 & 60.60 & 77.13 & 61.29 & 60.97 & 57.66 & 48.42 \\
    \textsc{AlignXplore+}       & 70.56 & 86.09 & 65.20 & 83.43 & 60.29 & 78.16 & 64.50 & 54.58 \\
    \bottomrule
  \end{tabular}}
\end{table*}

\begin{table*}[h]
  \centering
  \caption{Performance (\%) of different variants. Detailed results corresponding to Table~\ref{tab:ablation} in the main paper.}
  \label{tab:ablation-full}
  \adjustbox{max width=\linewidth}{
  \begin{tabular}{lcccccccc}
    \toprule
    \textbf{Model} & \textbf{MIND} & \textbf{Amazon} & \textbf{AlignX} &
    \textbf{MovieLens} & \textbf{Informativeness} & \textbf{Style} &
    \textbf{Expertise} & \textbf{PersonaMem} \\
    \midrule
    \midrule
    \multicolumn{9}{c}{\textbf{Full-history Preference Inference}} \\
    w/o RL           & 67.87 & 85.86 & 65.03 & 74.87 & 75.24 & 87.16 & 82.66 & 51.22 \\
    w/o SFT        & 68.26 & 84.88 & 64.66 & 74.66 & 77.90 & 86.16 & 84.33 & 53.00 \\
    w/o $R_{f}$ & 71.50 & 86.39 & 73.66 & 74.53 & 77.24 & 87.00 & 80.66 & 55.30 \\
    \textsc{AlignXplore+}      & 71.36 & 86.39 & 75.03 & 75.80 & 78.07 & 86.33 & 82.50 & 58.08 \\
    \midrule
    \midrule
    \multicolumn{9}{c}{\textbf{Streaming Preference Inference}} \\
    w/o RL          & 68.13 & 84.46 & 62.50 & 76.40 & 72.92 & 79.50 & 77.16 & 47.24 \\
    w/o SFT        & 66.86 & 84.78 & 66.46 & 76.93 & 75.58 & 80.66 & 79.50 & 49.82 \\
    w/o $R_{f}$ & 70.90 & 85.32 & 71.03 & 75.43 & 75.08 & 80.33 & 78.83 & 51.88 \\
    \textsc{AlignXplore+}       & 71.80 & 85.35 & 73.67 & 77.23 & 76.57 & 80.33 & 78.50 & 54.58 \\
    \bottomrule
  \end{tabular}}
\end{table*}

\begin{table}[t]
  \centering
  \caption{Performance (\%) of models trained with 10\% sub-datasets constructed using our method and random sampling (RDM). Detailed results corresponding to Table~\ref{tab:mini-random-sft-rl} in the appendix.}
  \label{tab:data_ablation}
  \adjustbox{max width=\linewidth}{
  \begin{tabular}{lcccccccc}
    \toprule
    \textbf{Method} & \textbf{MIND} & \textbf{Amazon} & \textbf{AlignX} &
    \textbf{MovieLens} & \textbf{Informativeness} & \textbf{Style} &
    \textbf{Expertise} & \textbf{PersonaMem} \\
    \midrule
    \midrule
    \multicolumn{9}{c}{\textbf{SFT-Only}} \\
    \midrule
    \multicolumn{9}{c}{\textbf{Full-history Preference Inference}} \\
    RDM              & 59.63 & 82.52 & 64.46 & 70.76 & 70.26 & 87.33 & 81.50 & 52.62 \\
    Ours                & 60.40 & 81.62 & 65.20 & 70.36 & 73.08 & 86.00 & 82.00 & 55.16 \\
    \midrule
    \multicolumn{9}{c}{\textbf{Streaming Preference Inference}} \\
    RDM              & 59.36 & 82.12 & 65.13 & 71.13 & 68.27 & 81.16 & 79.83 & 49.48 \\
    Ours                & 59.90 & 81.42 & 66.90 & 70.00 & 70.76 & 83.66 & 79.16 & 53.68 \\
    \midrule
    \midrule
    \multicolumn{9}{c}{\textbf{SFT \& RL}} \\
    \midrule
    \multicolumn{9}{c}{\textbf{Full-history Preference Inference}} \\
    RDM       & 59.63 & 81.59 & 65.43 & 69.00 & 53.18 & 75.08 & 85.00 & 81.83 \\
    Ours      & 60.70 & 82.65 & 67.83 & 69.23 & 55.28 & 76.91 & 87.66 & 84.50 \\
    \midrule
    \multicolumn{9}{c}{\textbf{Streaming Preference Inference}} \\
    RDM       & 59.13 & 81.32 & 65.67 & 69.73 & 50.10 & 67.61 & 83.50 & 79.83 \\
    Ours      & 59.60 & 81.78 & 68.40 & 70.03 & 52.52 & 74.75 & 85.16 & 80.00 \\

    \bottomrule
  \end{tabular}}
\end{table}

\end{document}